\documentclass[10pt,twocolumn,letterpaper]{article}

\usepackage{cvpr}
\usepackage{times}
\usepackage{epsfig}
\usepackage{graphicx}
\usepackage{amsmath}
\usepackage{amssymb}
\usepackage{color}
\usepackage{algorithm}
\usepackage{algorithmic}
\usepackage{multirow}
\usepackage{booktabs}
\usepackage{pifont}

\usepackage{tabulary,overpic,xcolor}
\usepackage[caption=false]{subfig}
\definecolor{citecolor}{RGB}{34,139,34}
\usepackage[pagebackref=true,breaklinks=true,letterpaper=true,colorlinks,
  citecolor=citecolor,bookmarks=false]{hyperref}

\newcommand{\cmark}{\ding{51}}
\newcommand{\xmark}{\ding{55}}

\newcommand{\bd}[1]{\textbf{#1}}
\newcommand{\app}{\raise.17ex\hbox{$\scriptstyle\sim$}}

\newcolumntype{x}[1]{>{\centering\arraybackslash}p{#1pt}}

\newlength\savewidth\newcommand\shline{\noalign{\global\savewidth\arrayrulewidth
  \global\arrayrulewidth 1pt}\hline\noalign{\global\arrayrulewidth\savewidth}}
\newcommand{\tablestyle}[2]{\setlength{\tabcolsep}{#1}\renewcommand{\arraystretch}{#2}\centering\footnotesize}
\makeatletter\renewcommand\paragraph{\@startsection{paragraph}{4}{\z@}
  {.5em \@plus1ex \@minus.2ex}{-.5em}{\normalfont\normalsize\bfseries}}\makeatother

\usepackage[pagebackref=true,breaklinks=true,letterpaper=true,colorlinks,bookmarks=false]{hyperref}

\cvprfinalcopy 

\ifcvprfinal\fi
\begin{document}

\title{Cascade R-CNN: Delving into High Quality Object Detection}

\author{Zhaowei Cai\\
UC San Diego\\
{\tt\small zwcai@ucsd.edu}
\and
Nuno Vasconcelos\\
UC San Diego\\
{\tt\small nuno@ucsd.edu}
}

\maketitle

\begin{abstract}
In object detection, an intersection over union (IoU) threshold is required to define positives and negatives. An object detector, trained with low IoU threshold, e.g. 0.5, usually produces noisy detections. However, detection performance tends to degrade with increasing the IoU thresholds. Two main factors are responsible for this: 1) overfitting during training, due to exponentially vanishing positive samples, and 2) inference-time mismatch between the IoUs for which the detector is optimal and those of the input hypotheses. A multi-stage object detection architecture, the Cascade R-CNN, is proposed to address these problems. It consists of a sequence of detectors trained with increasing IoU thresholds, to be sequentially more selective against close false positives. The detectors are trained stage by stage, leveraging the observation that the output of a detector is a good distribution for training the next higher quality detector. The resampling of progressively improved hypotheses guarantees that all detectors have a positive set of examples of equivalent size, reducing the overfitting problem. The same cascade procedure is applied at inference, enabling a closer match between the hypotheses and the detector quality of each stage. A simple implementation of the Cascade R-CNN is shown to surpass all single-model object detectors on the challenging COCO dataset. Experiments also show that the Cascade R-CNN is widely applicable across detector architectures, achieving consistent gains independently of the baseline detector strength. The code will be made available at https://github.com/zhaoweicai/cascade-rcnn.
\end{abstract}


\section{Introduction}
\label{sec:intro}

Object detection is a complex problem, requiring the solution of two main tasks. First, the detector must solve the {\it recognition\/} problem, to distinguish foreground objects from background and assign them the proper object class labels. Second, the detector must solve the {\it localization\/} problem, to assign accurate bounding boxes to different objects. Both of these are particularly difficult because the detector faces many ``close'' false positives, corresponding to ``close but not correct'' bounding boxes. The detector must find the true positives while suppressing these close false positives.

\begin{figure}[!t]
\begin{minipage}[b]{.495\linewidth}
\centering
\centerline{\epsfig{figure=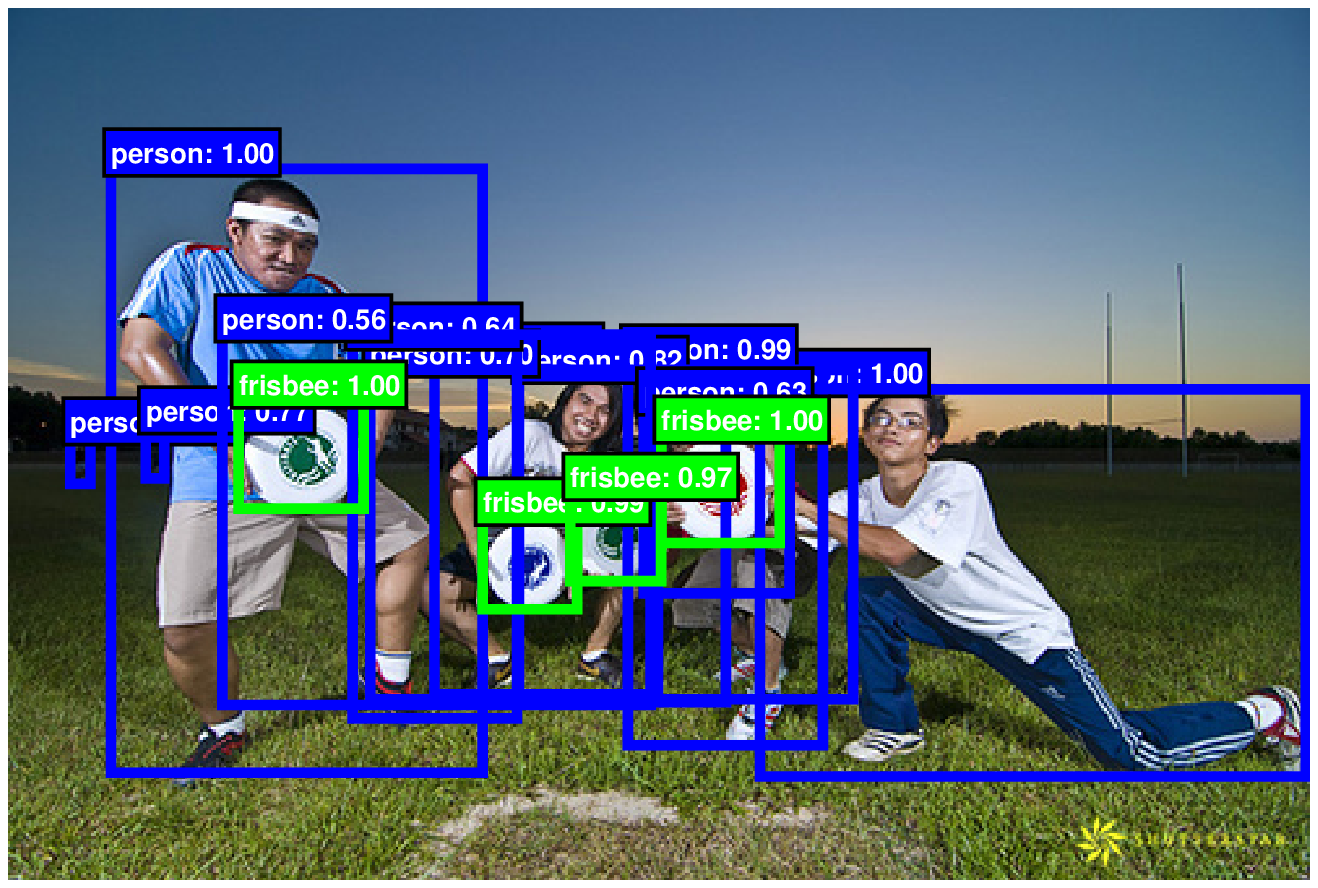,width=4.1cm,height=2.75cm}}{(a) Detection of $u=0.5$}
\end{minipage}
\hfill
\begin{minipage}[b]{.495\linewidth}
\centering
\centerline{\epsfig{figure=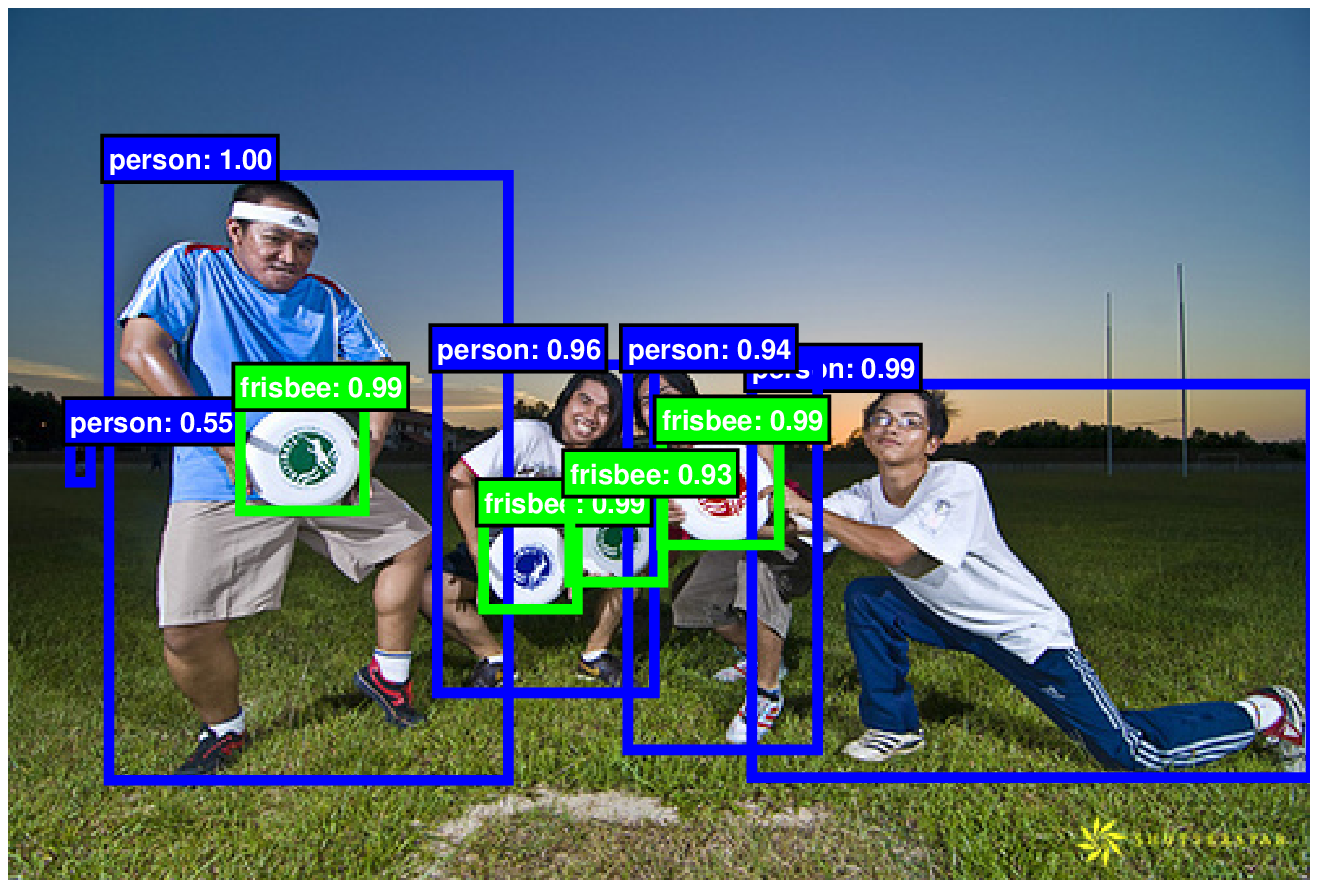,width=4.1cm,height=2.75cm}}{(b) Detection of $u=0.7$}
\end{minipage}\\
\hfill
\begin{minipage}[b]{.48\linewidth}
\centering
\centerline{\epsfig{figure=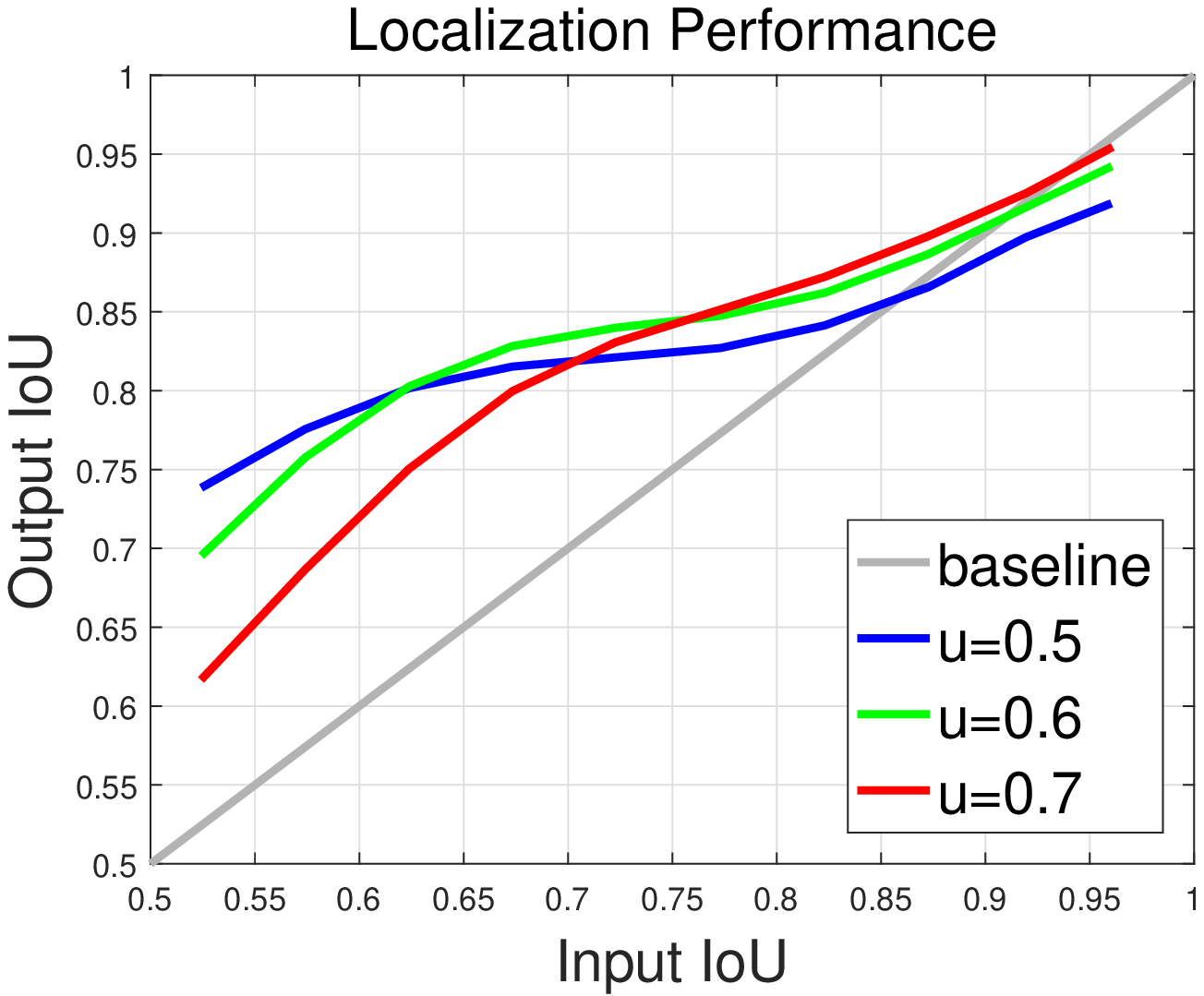,width=4.4cm,height=3.3cm}}{(c) Regressor}
\end{minipage}
\hfill
\begin{minipage}[b]{.48\linewidth}
\centering
\centerline{\epsfig{figure=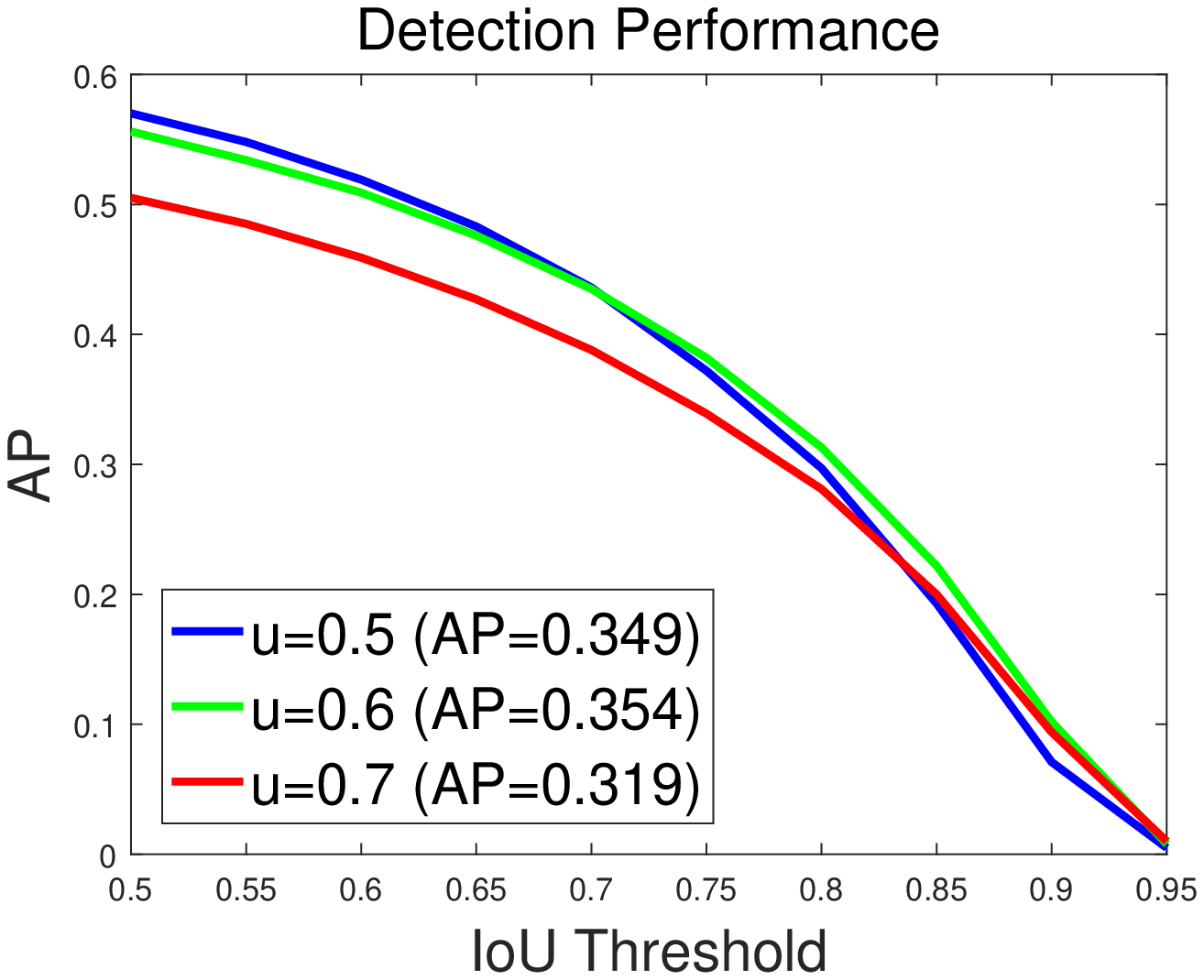,width=4.4cm,height=3.3cm}}{(d) Detector}
\end{minipage}
\caption{The detection outputs, localization and detection performance of object detectors of increasing IoU threshold $u$.}
\label{fig:motivation}
\end{figure}

Many of the recently proposed object detectors are based on the
two-stage R-CNN framework \cite{DBLP:conf/cvpr/GirshickDDM14,DBLP:conf/iccv/Girshick15,DBLP:conf/nips/RenHGS15,lin2017feature}, where detection is framed
as a multi-task learning problem that combines classification and bounding
box regression. Unlike object recognition, an intersection over union (IoU)
threshold is required to define positives/negatives. However, the commonly used threshold values $u$, typically $u=0.5$, establish quite a loose
requirement for positives. The resulting detectors frequently produce noisy
bounding boxes, as shown in Figure \ref{fig:motivation} (a). Hypotheses
that most humans would consider close false positives frequently pass the $IoU \geq 0.5$ test. While the examples assembled under the $u=0.5$ criterion are
rich and diversified, they make it difficult to train detectors that can
effectively reject close false positives.

In this work, we define the {\it quality\/} of an hypothesis as its IoU with
the ground truth, and the {\it quality of the detector\/} as the IoU threshold
$u$ used to train it. The goal is to investigate the, so far, poorly
researched problem of learning high quality object detectors,
whose outputs contain few close false positives, as shown in
Figure \ref{fig:motivation} (b). The basic idea is that a single detector \
can only be optimal for a single quality level. This is known in the
cost-sensitive learning literature~\cite{DBLP:conf/ijcai/Elkan01,
DBLP:journals/pami/Masnadi-ShiraziV11}, where the optimization of
different points of the receiver operating characteristic (ROC) requires
different loss functions. The main difference is that we consider
the optimization for a given IoU threshold, rather than false positive rate.

The idea is illustrated by Figure \ref{fig:motivation} (c) and (d), which
present the localization and detection performance, respectively, of
three detectors trained with IoU thresholds of $u=0.5,0.6,0.7$. The
localization performance is evaluated
as a function of the IoU of the input proposals, and the detection
performance as a function of IoU threshold, as in
COCO \cite{DBLP:conf/eccv/LinMBHPRDZ14}. Note that, in
Figure \ref{fig:motivation} (c), each bounding box regressor performs
best for examples of IoU close to the threshold that the detector
was trained. This also holds for detection performance, up to overfitting.
Figure \ref{fig:motivation} (d) shows that, the detector of $u=0.5$
outperforms the detector of $u=0.6$ for low IoU examples, underperforming
it at higher IoU levels. In general, a detector optimized at a single IoU
level is not necessarily optimal at other levels.
These observations suggest that higher quality detection requires a
closer quality {\it match\/} between the detector and the
hypotheses that it processes. In general, a detector can only
have high quality if presented with high quality proposals.

However, to produce a high quality detector, it does not suffice to
simply increase $u$ during training. In fact, as seen for the detector
of $u=0.7$ of Figure \ref{fig:motivation} (d), this can degrade detection
performance. The problem is that the distribution of hypotheses
out of a proposal detector is usually heavily imbalanced towards low
quality. In general, forcing larger IoU thresholds leads to an
exponentially smaller numbers of positive training samples. This is
particularly problematic for neural networks, which are known to be very
example intensive, and makes the ``high $u$'' training strategy
quite prone to overfitting. Another difficulty is the
mismatch between the quality of the detector and that of the testing
hypotheses at inference. As shown in Figure \ref{fig:motivation},
high quality detectors are only necessarily optimal for high quality
hypotheses. The detection could be suboptimal when they are asked to work
on the hypotheses of other quality levels.

In this paper, we propose a new detector architecture, Cascade R-CNN,
that addresses these problems. It is a multi-stage extension of the
R-CNN, where detector stages deeper into the cascade are sequentially more
selective against close false positives. The cascade of R-CNN stages are
trained sequentially, using the output of one stage to train the next. This
is motivated by the observation that the output IoU of a regressor is
almost invariably better than the input IoU. This observation can be made
in Figure \ref{fig:motivation} (c), where all plots are above the gray
line. It suggests that the output of a detector trained with a certain
IoU threshold is a good distribution to train the detector of the next
higher IoU threshold. This is similar to {\it boostrapping\/} methods
commonly used to assemble datasets in object detection
literature \cite{DBLP:journals/ijcv/ViolaJ04,DBLP:journals/pami/FelzenszwalbGMR10}. The main difference is that the resampling procedure of the
Cascade R-CNN does not aim to mine hard negatives. Instead, by adjusting
bounding boxes, each stage aims to find a good set of close false positives
for training the next stage. When operating in this manner, a sequence of
detectors adapted to increasingly higher IoUs can beat the overfitting
problem, and thus be effectively trained. At inference, the same cascade
procedure is applied. The progressively improved hypotheses are better
matched to the increasing detector quality at each stage. This enables
higher detection accuracies, as suggested by Figure \ref{fig:motivation} (c)
and (d).

The Cascade R-CNN is quite simple to implement and trained end-to-end.
Our results show that a vanilla implementation, without any bells and whistles,
surpasses all previous state-of-the-art \emph{single-model} detectors by a
large margin, on the challenging COCO detection
task \cite{DBLP:conf/eccv/LinMBHPRDZ14}, especially under the higher quality
evaluation metrics. In addition, the Cascade R-CNN can be built with any
two-stage object detector based on the R-CNN framework. We have
observed consistent gains (of 2$\sim$4 points), at a marginal increase in
computation. This gain is independent of the strength of the baseline object
detectors. We thus believe that this simple and effective detection
architecture can be of interest for many object detection research efforts.

\section{Related Work}

Due to the success of  the R-CNN \cite{DBLP:conf/cvpr/GirshickDDM14}
architecture, the two-stage formulation of the detection problems,
by combining a proposal detector and a region-wise classifier has
become predominant in the recent past. To reduce redundant CNN
computations in the R-CNN, the SPP-Net \cite{DBLP:conf/eccv/HeZR014} and
Fast-RCNN \cite{DBLP:conf/iccv/Girshick15} introduced
the idea of region-wise feature extraction, significantly speeding up
the overall detector. Later, the Faster-RCNN \cite{DBLP:conf/nips/RenHGS15} achieved  further speeds-up by introducing a Region Proposal
Network (RPN). This architecture has become a leading object detection
framework. Some more recent works have extended it to address various
problems of detail. For example, the R-FCN \cite{DBLP:conf/nips/DaiLHS16}
proposed efficient region-wise fully convolutions without accuracy loss,
to avoid the heavy region-wise CNN computations of the Faster-RCNN;
while the MS-CNN \cite{DBLP:conf/eccv/CaiFFV16} and FPN \cite{lin2017feature}
detect proposals at multiple output layers, so as to alleviate the scale
mismatch between the RPN receptive fields and actual object size, for
high-recall proposal detection.

Alternatively, one-stage object detection architectures have also become popular, mostly due to their computational efficiency. These architectures are
close to the classic sliding window strategy \cite{DBLP:journals/ijcv/ViolaJ04,
DBLP:journals/pami/FelzenszwalbGMR10}. YOLO \cite{DBLP:conf/cvpr/RedmonDGF16}
outputs very sparse detection results by forwarding the input image once.
When implemented with an efficient backbone network, it enables real time
object detection with fair performance. SSD \cite{DBLP:conf/eccv/LiuAESRFB16}
detects objects in a way similar to the RPN \cite{DBLP:conf/nips/RenHGS15},
but uses multiple feature maps at different resolutions to cover objects at
various scales. The main limitation of these architectures is that their
accuracies are typically below that of two-stage detectors. Recently,
RetinaNet \cite{lin2017focal} was proposed to address the extreme
foreground-background class imbalance in dense object detection,
achieving better results than state-of-the-art two-stage object detectors.

Some explorations in multi-stage object detection have also been proposed.
The multi-region detector \cite{DBLP:conf/iccv/GidarisK15}
introduced \textit{iterative bounding box regression}, where a
R-CNN is applied several times, to produce better bounding boxes.
CRAFT \cite{DBLP:conf/cvpr/YangYLL16} and
AttractioNet \cite{DBLP:conf/bmvc/GidarisK16} used a multi-stage procedure
to generate accurate proposals, and forwarded them to a
Fast-RCNN. \cite{DBLP:conf/cvpr/LiLSBH15,DBLP:journals/corr/OuyangWZW17}
embedded the classic cascade architecture
of \cite{DBLP:journals/ijcv/ViolaJ04} in object detection networks.
\cite{DBLP:conf/cvpr/DaiHS16} iterated a detection and a segmentation task
alternatively, for instance segmentation.

\section{Object Detection}

In this paper, we extend the two-stage architecture of the
Faster-RCNN \cite{DBLP:conf/nips/RenHGS15,lin2017feature}, shown in
Figure \ref{fig:framework} (a). The first stage is a proposal sub-network
(``H0''), applied to the entire image, to produce preliminary detection
hypotheses, known as object proposals. In the second stage, these
hypotheses are then processed by a region-of-interest detection
sub-network (``H1''), denoted as detection head. A final classification
score (``C'') and a bounding box (``B'') are assigned to each hypothesis.
We focus on modeling a multi-stage detection sub-network, and adopt, but
are not limited to, the RPN \cite{DBLP:conf/nips/RenHGS15} for proposal
detection.

\subsection{Bounding Box Regression}
\label{subsec:bbox}

A bounding box $\textbf{b}=(b_x,b_y,b_w,b_h)$ contains the four coordinates of
an image patch $x$. The task of bounding box
regression is to regress a candidate bounding box $\textbf{b}$ into a target
bounding box $\textbf{g}$, using a regressor $f(x,\textbf{b})$. This is
learned from a training sample
$\{\textbf{g}_i,\textbf{b}_i\}$, so as to minimize the bounding box risk
\begin{equation}
  {\cal R}_{loc}[f] = \sum_{i=1}^{N}L_{loc}(f(x_i,\textbf{b}_i),\textbf{g}_i),
  \label{eq:bbrisk}
\end{equation}
where $L_{loc}$ was a $L_2$ loss function in R-CNN \cite{DBLP:conf/cvpr/GirshickDDM14}, but updated to a smoothed $L_1$ loss function in Fast-RCNN \cite{DBLP:conf/iccv/Girshick15}. To encourage a regression invariant to scale and location, $L_{loc}$ operates on the distance vector $\Delta=(\delta_x,\delta_y,\delta_w,\delta_h)$ defined by
\begin{equation}
\label{equ:delta}
\begin{array}{cl}\delta_x=(g_x-b_x)/b_w,\quad\delta_y=(g_y-b_y)/b_h\\
\delta_w=\log(g_w/b_w),\quad\delta_h=\log(g_h/b_h).\end{array}
\end{equation}
Since bounding box regression usually performs minor adjustments on $b$, the numerical values of (\ref{equ:delta}) can be very
small. Hence, the risk of \eqref{eq:bbrisk} is usually much smaller
than the classification risk. To improve the effectiveness of multi-task
learning, $\Delta$ is usually normalized by its mean and variance, i.e.
$\delta_x$ is replaced by $\delta_x'=(\delta_x-\mu_x)/\sigma_x$. This is widely
used in the literature \cite{DBLP:conf/nips/RenHGS15,DBLP:conf/eccv/CaiFFV16,DBLP:conf/nips/DaiLHS16,lin2017feature,he2017mask}.

\begin{figure}[!t]
\begin{minipage}[b]{.32\linewidth}
\centering
\centerline{\epsfig{figure=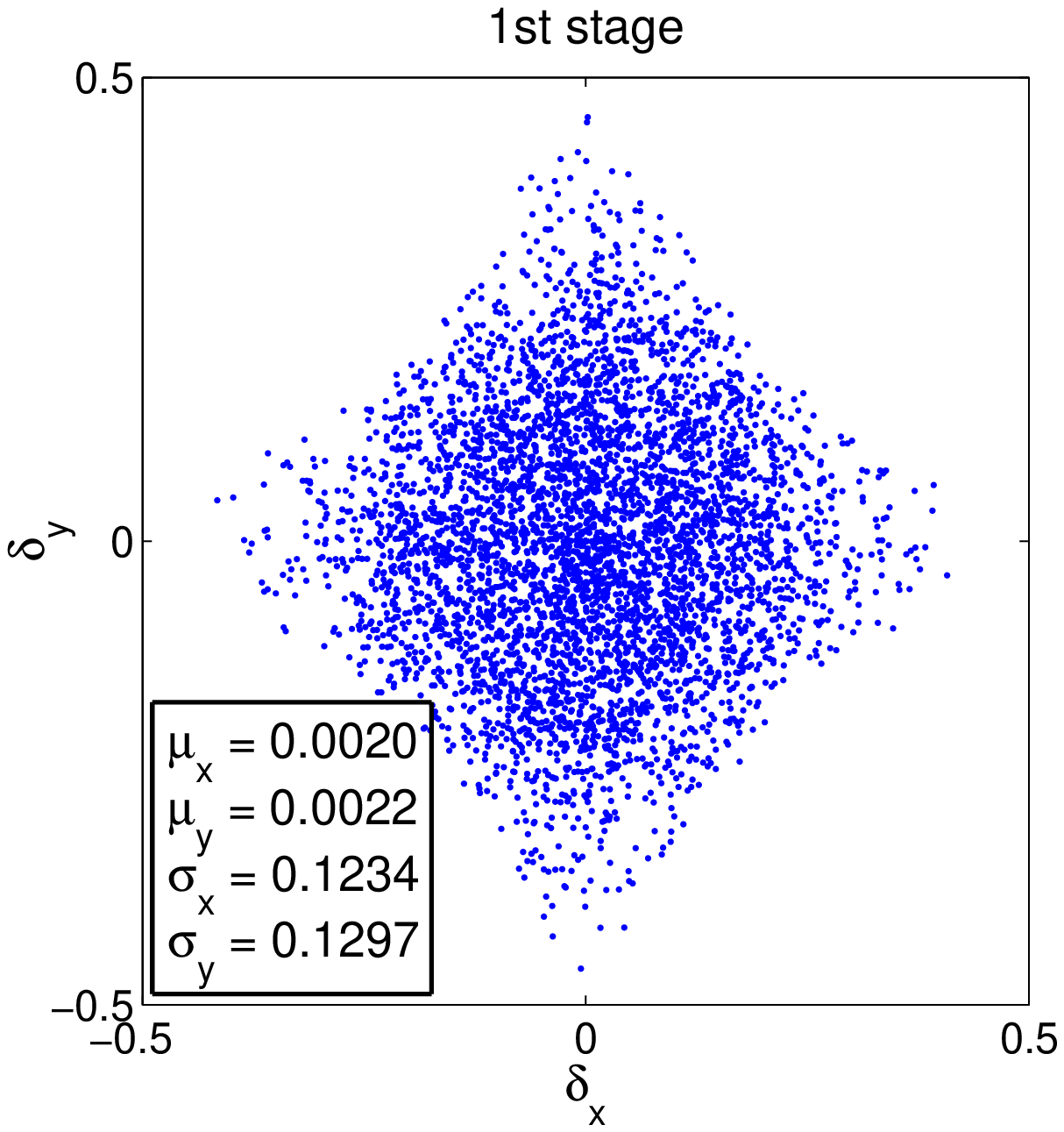,width=3cm,height=2.8cm}}
\end{minipage}
\hfill
\begin{minipage}[b]{.32\linewidth}
\centering
\centerline{\epsfig{figure=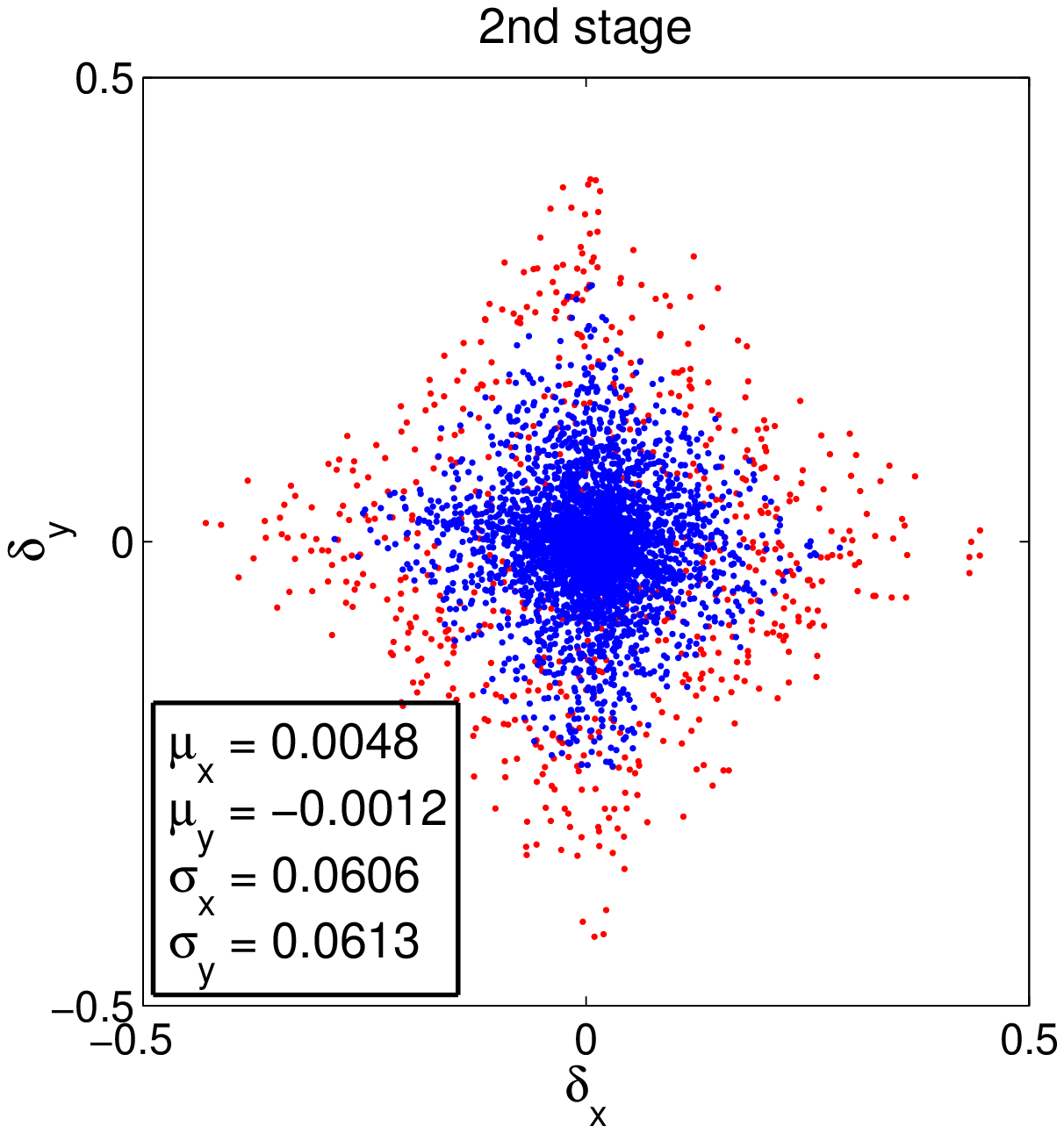,width=3cm,height=2.8cm}}
\end{minipage}
\hfill
\begin{minipage}[b]{.32\linewidth}
\centering
\centerline{\epsfig{figure=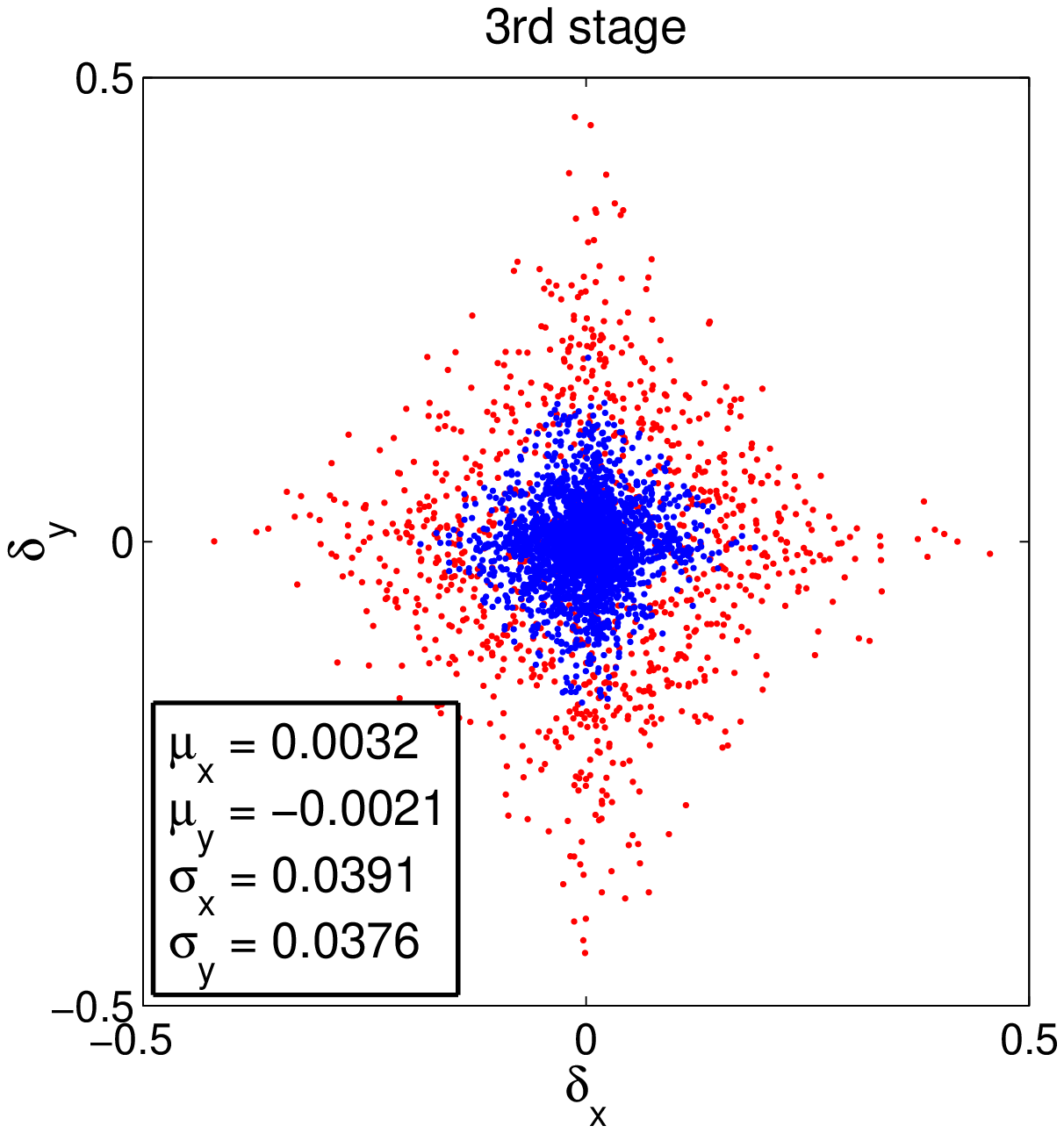,width=3cm,height=2.8cm}}
\end{minipage}\\
\hfill
\begin{minipage}[b]{.32\linewidth}
\centering
\centerline{\epsfig{figure=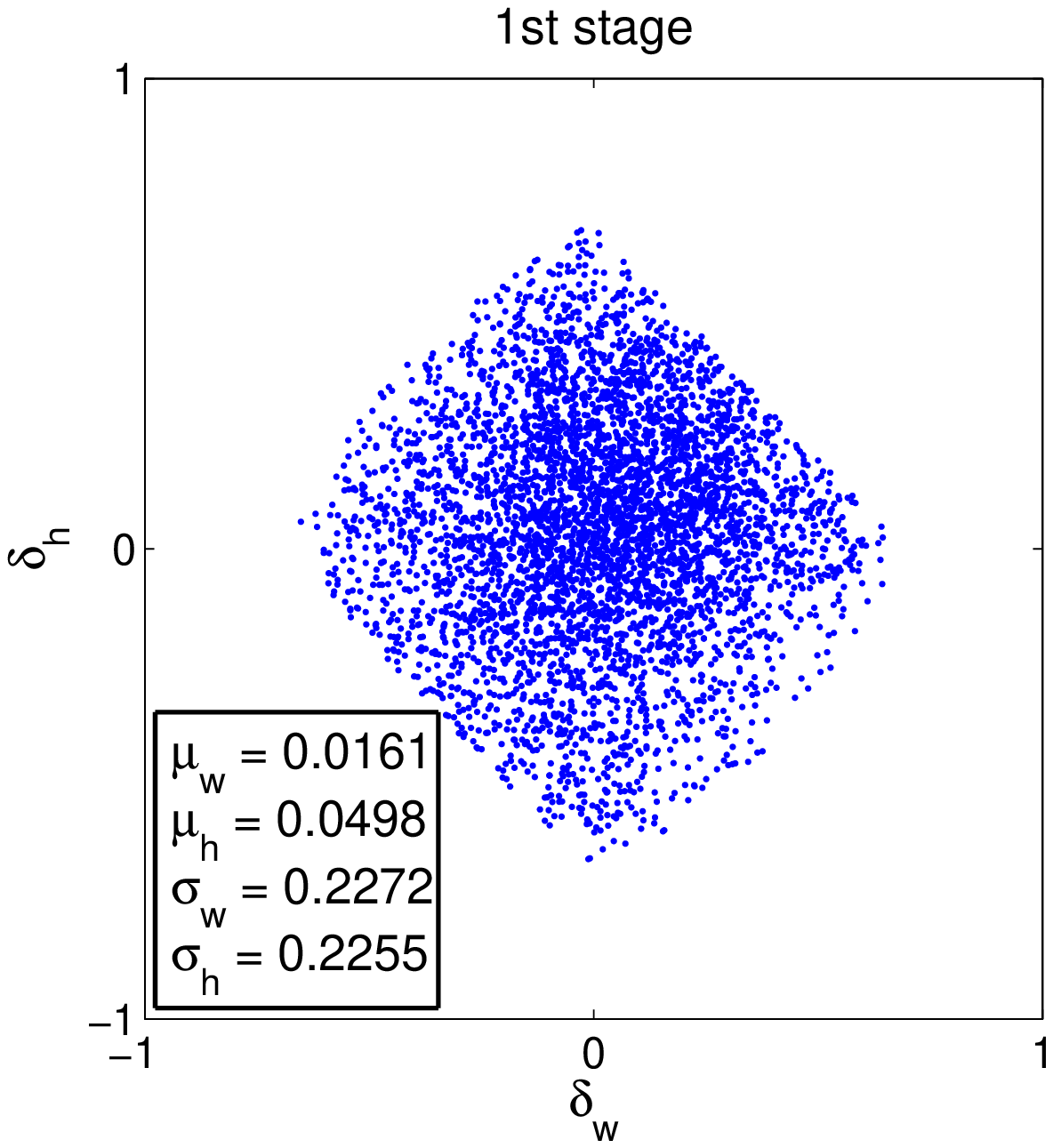,width=3cm,height=2.8cm}}
\end{minipage}
\hfill
\begin{minipage}[b]{.32\linewidth}
\centering
\centerline{\epsfig{figure=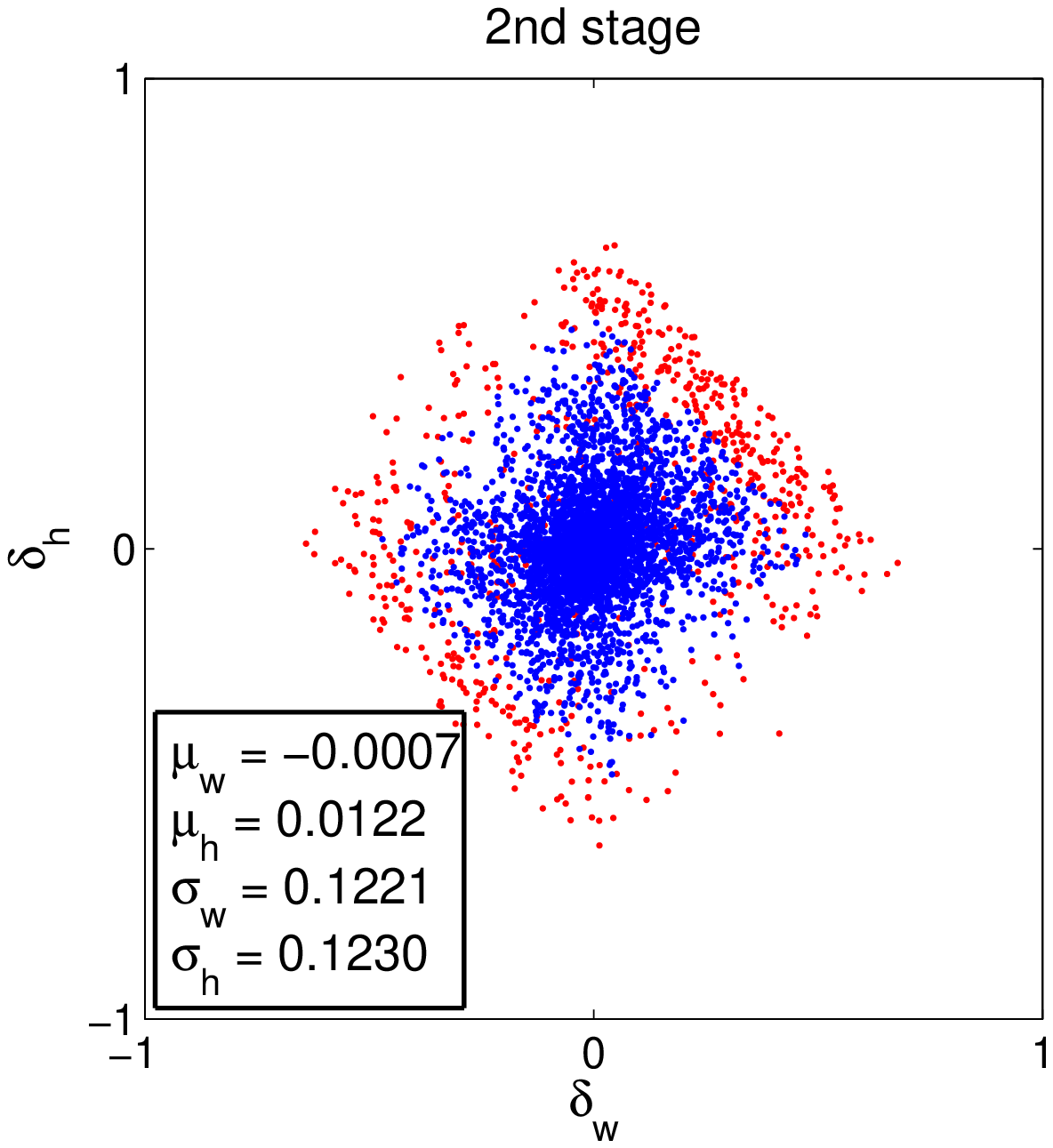,width=3cm,height=2.8cm}}
\end{minipage}
\hfill
\begin{minipage}[b]{.32\linewidth}
\centering
\centerline{\epsfig{figure=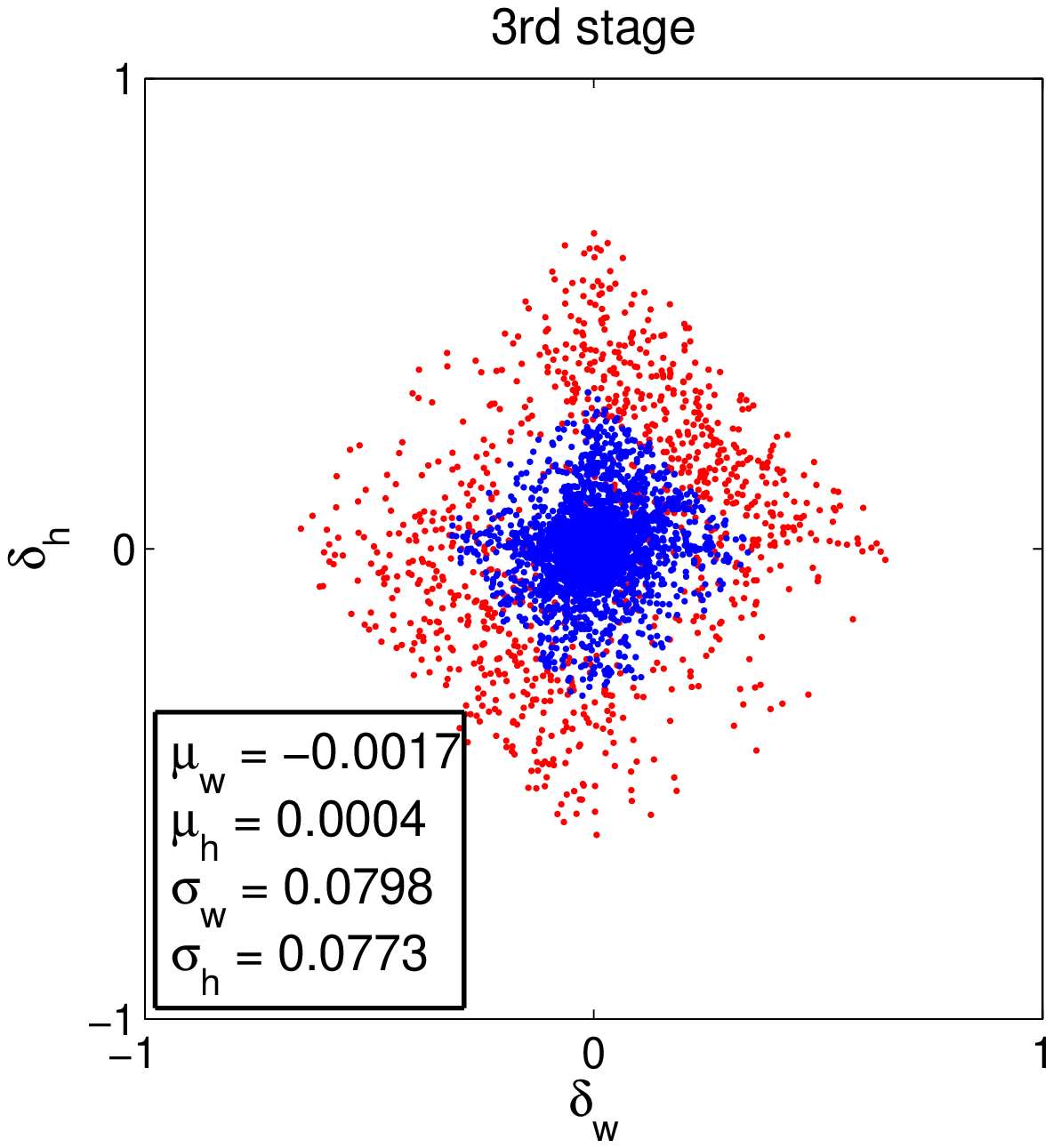,width=3cm,height=2.8cm}}
\end{minipage}
\caption{Sequential $\Delta$ distribution (without normalization) at different cascade stage. Red dots are outliers when using increasing IoU thresholds, and the statistics are obtained after outlier removal.}
\label{fig:distribution}
\end{figure}

\begin{figure*}[!t]
\begin{minipage}[b]{.17\linewidth}
\centering
\centerline{\epsfig{figure=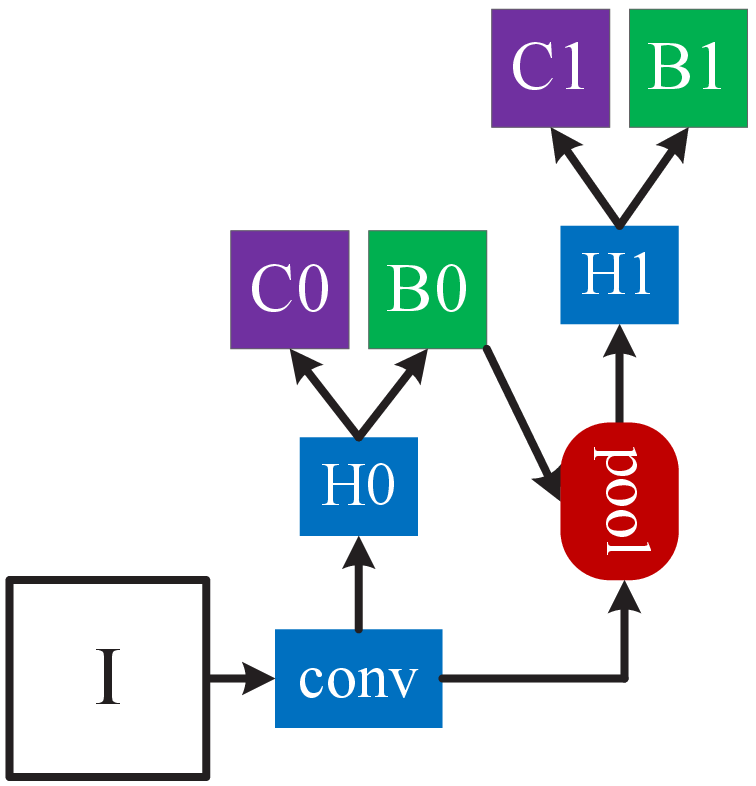,width=2.95cm,height=3.06cm}}{(a) Faster R-CNN}
\end{minipage}
\hfill
\begin{minipage}[b]{.3\linewidth}
\centering
\centerline{\epsfig{figure=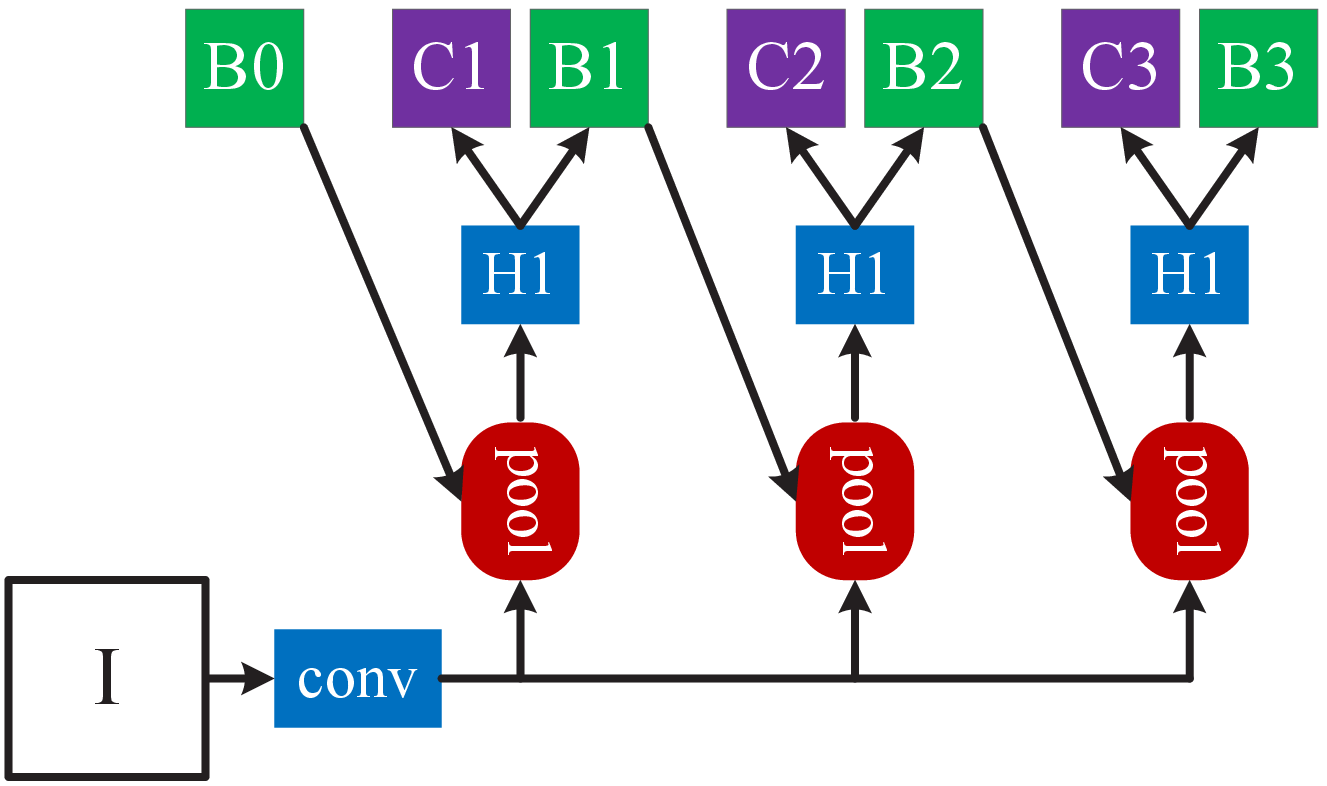,width=5.24cm,height=3.06cm}}{(b) Iterative BBox at inference}
\end{minipage}
\hfill
\begin{minipage}[b]{.21\linewidth}
\centering
\centerline{\epsfig{figure=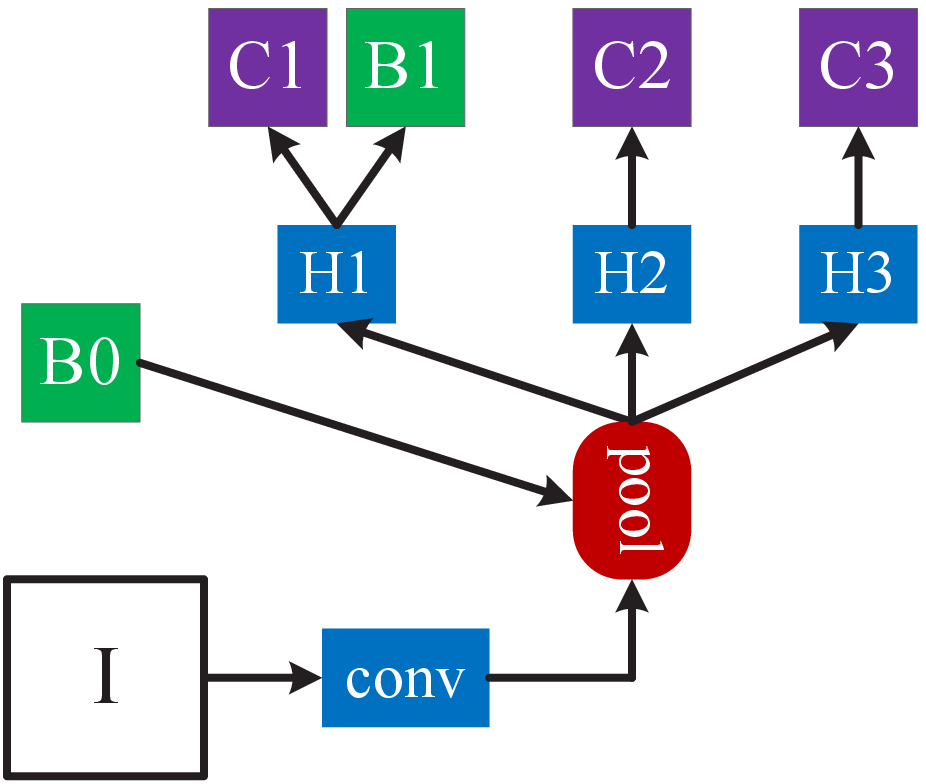,width=3.65cm,height=3.06cm}}{(c) Integral Loss}
\end{minipage}
\hfill
\begin{minipage}[b]{.3\linewidth}
\centering
\centerline{\epsfig{figure=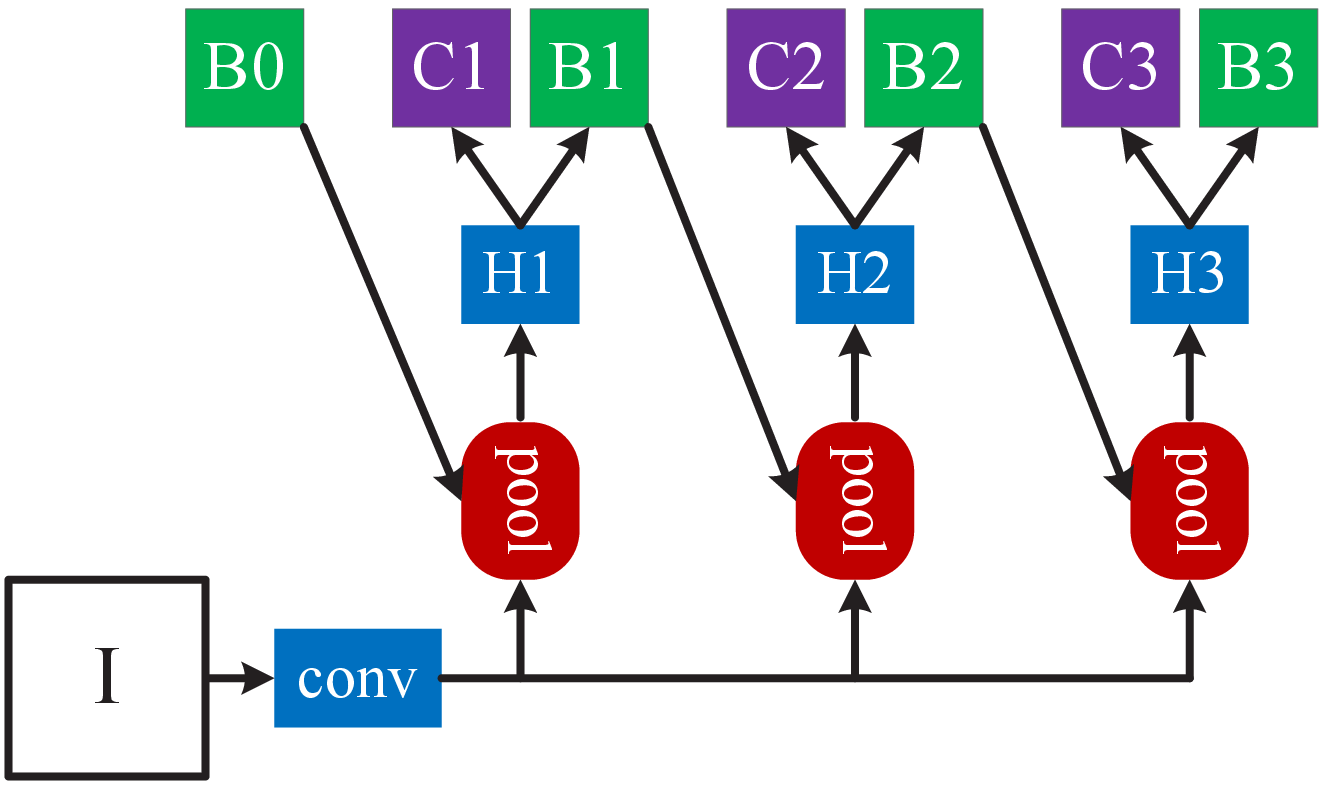,width=5.24cm,height=3.06cm}}{(d) Cascade R-CNN}
\end{minipage}
\caption{The architectures of different frameworks. ``I'' is input image, ``conv'' backbone convolutions, ``pool'' region-wise feature extraction, ``H'' network head, ``B'' bounding box, and ``C'' classification. ``B0'' is proposals in all architectures.}
\label{fig:framework}
\end{figure*}

Some works \cite{DBLP:conf/iccv/GidarisK15,DBLP:conf/bmvc/GidarisK16,DBLP:conf/cvpr/HeZRS16} have argued that a single regression step of $f$ is insufficient
for accurate localization. Instead, $f$ is applied iteratively, as a
post-processing step
\begin{equation}
\label{equ:iterative bbox}
f'(x,\textbf{b})=f\circ{f}\circ\cdots\circ{f}(x,\textbf{b}),
\end{equation}
to refine a bounding box $\textbf{b}$. This is called \textit{iterative bounding
box regression}, denoted as \textit{iterative BBox}. It can be implemented
with the inference architecture of Figure \ref{fig:framework} (b) where all
heads are the same. This idea, however, ignores two problems. First, as
shown in Figure \ref{fig:motivation}, a regressor $f$ trained at $u=0.5$, is
suboptimal for hypotheses of higher IoUs. It actually {\it degrades\/}
bounding boxes of IoU larger than $0.85$. Second, as shown in
Figure \ref{fig:distribution}, the distribution of bounding boxes changes
significantly after each iteration. While the regressor is optimal for the
initial distribution it can be quite suboptimal after that. Due to these
problems, \textit{iterative BBox} requires a fair amount of human
engineering, in the form of proposal accumulation, box voting, etc. \cite{DBLP:conf/iccv/GidarisK15,DBLP:conf/bmvc/GidarisK16,DBLP:conf/cvpr/HeZRS16},
and has somewhat unreliable gains. Usually, there is no benefit beyond
applying $f$ twice.

\subsection{Classification}
\label{subsec:classification}

The classifier is a function $h(x)$ that assigns an image patch $x$ to one
of $M+1$ classes, where class $0$ contains background and the remaining
the objects to detect. $h(x)$ is a $M+1$-dimensional estimate
of the posterior distribution over classes, i.e. $h_k(x)=p(y=k|x)$,
where $y$ is the class label. Given a training set $(x_i, y_i)$, it is
learned by minimizing a classification risk
\begin{equation}
  {\cal R}_{cls}[h] = \sum_{i=1}^{N}L_{cls}(h(x_i),y_i),
  \label{eq:clsrisk}
\end{equation}
where $L_{cls}$ is the classic cross-entropy loss.

\subsection{Detection Quality}

Since a bounding box usually includes an object and some amount of background, it is difficult to determine if a detection is positive or negative. This is usually addressed by the IoU metric. If the IoU is above a threshold $u$, the patch is considered an example of the class. Thus, the class label of a hypothesis $x$ is a function of $u$,
\begin{equation}
\label{equ:cls label}
y=\left\{
\begin{array}{cl}g_y, &\quad\textrm{$IoU(x,g)\geq{u}$}\\
0, &\quad\textrm{otherwise} \end{array}\right.
\end{equation}
where $g_y$ is the class label of the ground truth object $g$. This IoU threshold $u$ defines the quality of a detector.

Object detection is challenging because, no matter threshold, the detection
setting is highly adversarial. When $u$ is high, the positives contain
less background, but it is difficult to assemble enough positive training
examples. When $u$ is low, a richer and more diversified positive training
set is available, but the trained detector has little incentive to reject
close false positives. In general, it is very difficult to ask a single
classifier to perform uniformly well over all IoU levels. At inference,
since the majority of the hypotheses produced by a proposal detector,
e.g. RPN \cite{DBLP:conf/nips/RenHGS15} or selective
search \cite{DBLP:journals/ijcv/UijlingsSGS13}, have low quality, the detector
must be more discriminant for lower quality hypotheses. A standard
compromise between these conflicting requirements is to settle on $u=0.5$.
This, however, is a relatively low threshold, leading to low quality
detections that most humans consider close false positives, as shown in
Figure \ref{fig:motivation} (a).

\begin{figure}[!t]
\begin{minipage}[b]{.3\linewidth}
\centering
\centerline{\epsfig{figure=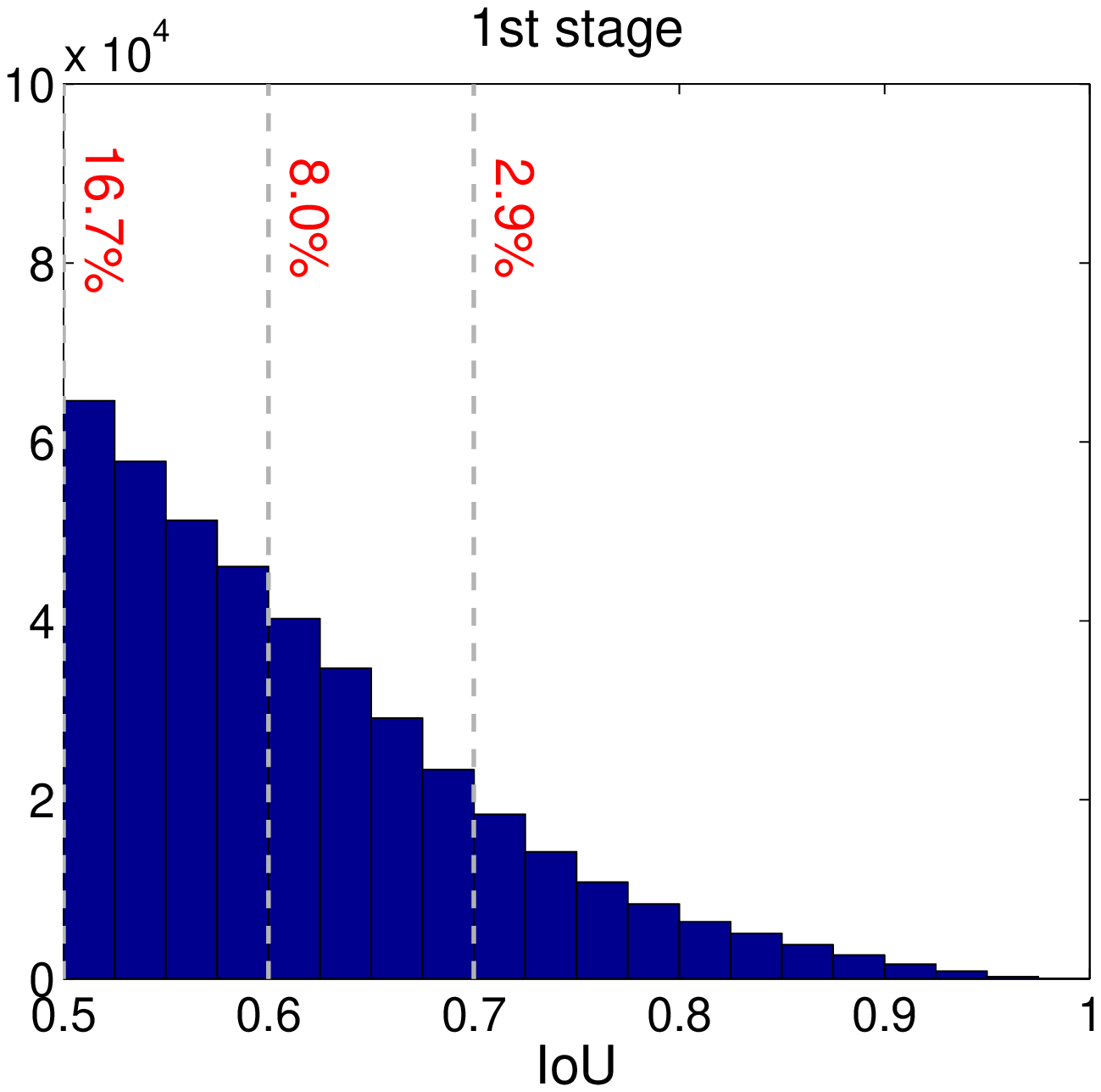,width=3.2cm,height=2.6cm}}
\end{minipage}
\hfill
\begin{minipage}[b]{.3\linewidth}
\centering
\centerline{\epsfig{figure=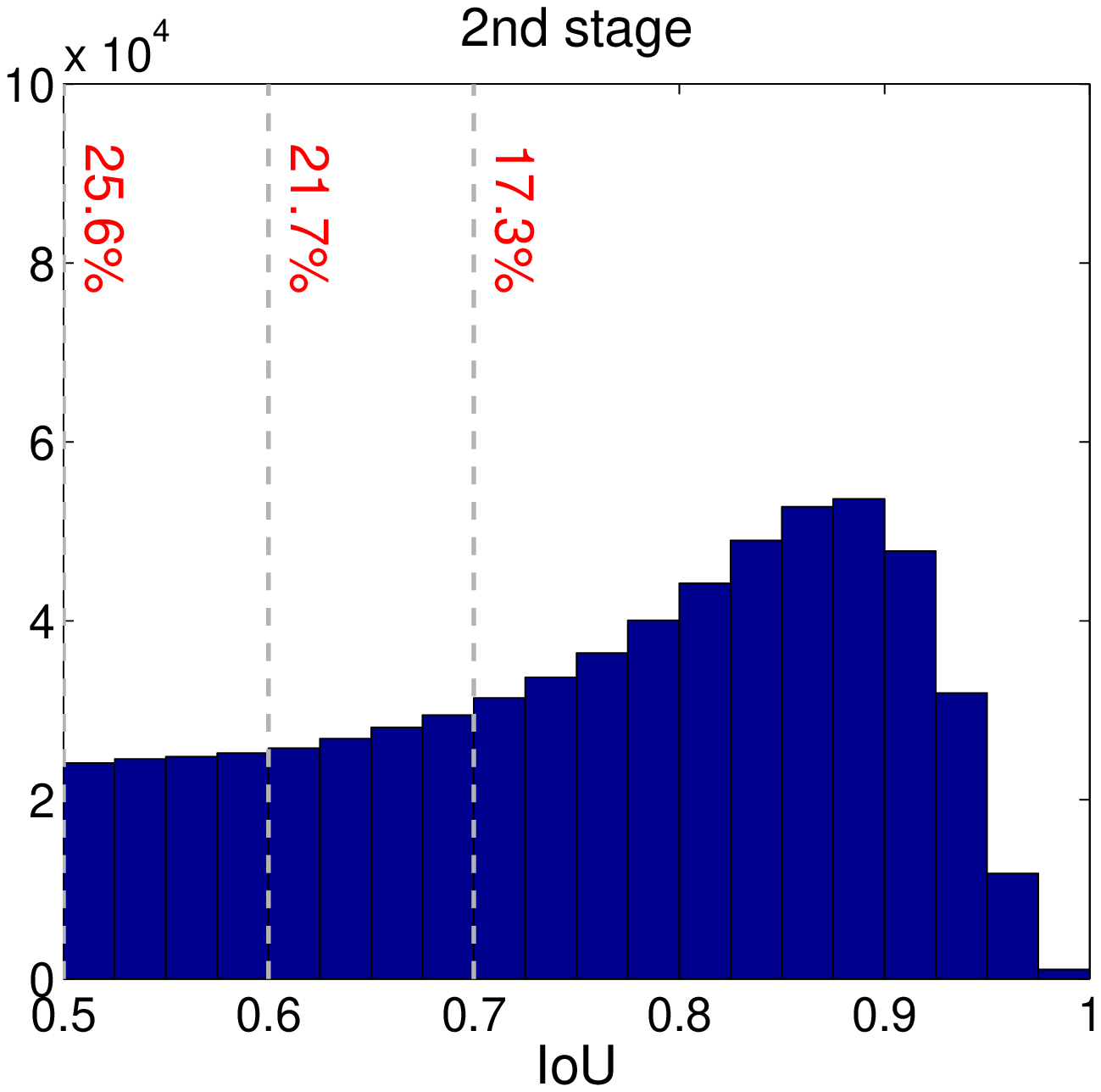,width=3.2cm,height=2.6cm}}
\end{minipage}
\hfill
\begin{minipage}[b]{.3\linewidth}
\centering
\centerline{\epsfig{figure=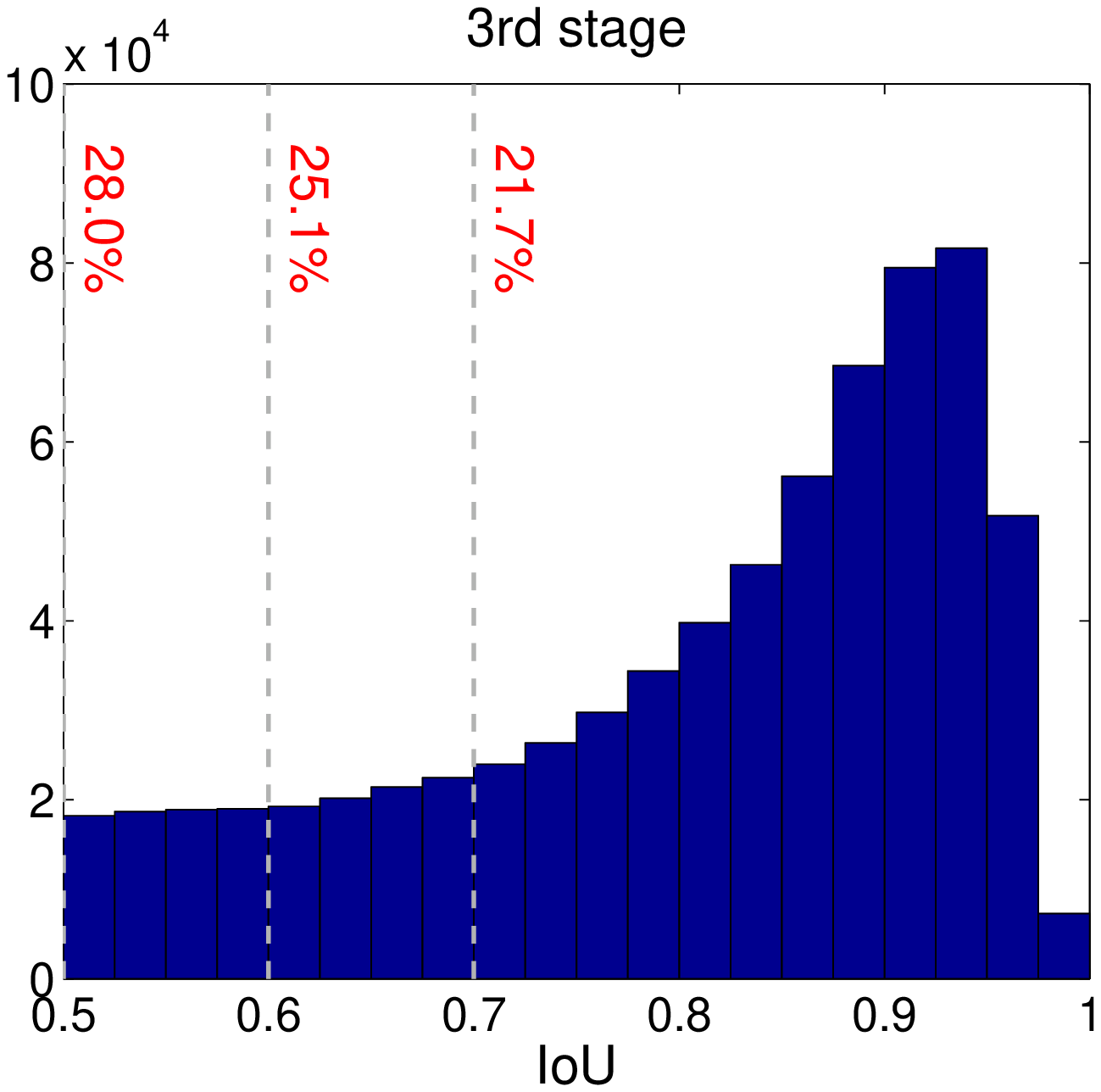,width=3.2cm,height=2.6cm}}
\end{minipage}
\caption{The IoU histogram of training samples. The distribution at 1st stage is the output of RPN. The red numbers are the positive percentage higher than the corresponding IoU threshold.}
\label{fig:hist}
\end{figure}

A na\"ive solution is to develop an ensemble of classifiers, with the
architecture of Figure \ref{fig:framework} (c), optimized with
a loss that targets various quality levels,
\begin{equation}
  \label{equ:intloss}
  L_{cls}(h(x),y)=\sum_{u\in{U}}L_{cls}(h_u(x),y_u),
\end{equation}
where $U$ is a set of IoU thresholds. This is closely related
to the \textit{integral loss} of \cite{DBLP:conf/bmvc/ZagoruykoLLPGCD16}, in
which $U=\{0.5,0.55,\cdots,0.75\}$, designed to fit the evaluation
metric of the COCO challenge. By definition, the classifiers need to be
ensembled at inference. This solution fails to address the problem that
the different losses of \eqref{equ:intloss} operate on different numbers of
positives. As shown in the first figure of Figure \ref{fig:hist}, the set of
positive samples decreases quickly with $u$. This is particularly
problematic because the high quality classifiers are prone to overfitting.
In addition, those high quality classifiers are required to process
proposals of overwhelming low quality at inference, for which they are not
optimized. Due to all this, the ensemble of (\ref{equ:intloss})
fails to achieve higher accuracy at most quality levels, and the architecture
has very little gain over that of Figure \ref{fig:framework} (a).

\section{Cascade R-CNN}

In this section we introduce the proposed Cascade R-CNN object detection architecture of Figure \ref{fig:framework} (d).

\subsection{Cascaded Bounding Box Regression}
\label{subsec:cascade bbox}

As seen in Figure \ref{fig:motivation} (c), it is very difficult to ask
a single regressor to perform perfectly uniformly at all quality levels.
The difficult regression task can be decomposed into a sequence of simpler
steps, inspired by the works of cascade pose
regression \cite{DBLP:conf/cvpr/DollarWP10} and face
alignment \cite{DBLP:conf/cvpr/CaoWWS12,yan2013learn}. In the
Cascade R-CNN, it is framed as a cascaded regression problem, with the
architecture of Figure \ref{fig:framework} (d). This relies on a cascade
of {\it specialized\/} regressors
\begin{equation}
  f(x,\textbf{b})=f_T\circ{f}_{T-1}\circ\cdots\circ{f}_1(x,\textbf{b}),
\end{equation}
where $T$ is the total number of cascade stages. Note that each regressor
$f_t$ in the cascade is optimized \textit{w.r.t.} the sample
distribution $\{\textbf{b}^t\}$ arriving at the corresponding stage,
instead of the initial distribution of $\{\textbf{b}^1\}$. This cascade
improves hypotheses progressively.

It differs from the \textit{iterative BBox} architecture of
Figure \ref{fig:framework} (b) in several ways. First,
while \textit{iterative BBox} is a post-processing procedure used to
improve bounding boxes, cascaded regression is a {\it resampling\/}
procedure that changes the distribution of hypotheses to be processed
by the different
stages. Second, because it is used at both training and inference, there is
no discrepancy between training and inference distributions. Third,
the multiple specialized regressors $\{f_T,f_{T-1},\cdots,f_1\}$ are
optimized for the {\it resampled distributions\/} of the different stages.
This opposes to the single $f$ of (\ref{equ:iterative bbox}),
which is only optimal for the initial distribution. These differences enable
more precise localization than \textit{iterative BBox}, with no further
human engineering.

As discussed in Section \ref{subsec:bbox}, $\Delta=(\delta_x,\delta_y,\delta_w,\delta_h)$ in (\ref{equ:delta}) needs to be normalized by its mean and variance for effective multi-task learning. After each regression stage, these statistics will evolve sequentially, as displayed in Figure \ref{fig:distribution}. At training, the corresponding statistics are used to normalize $\Delta$ at each stage.

\subsection{Cascaded Detection}
\label{subsec:cascade}

As shown in the left of Figure \ref{fig:hist}, the distribution of
the initial hypotheses, e.g. RPN proposals, is heavily tilted towards low
quality. This inevitably induces ineffective learning of higher quality
classifiers. The Cascade R-CNN addresses the problem by
relying on cascade regression as a {\it resampling mechanism\/}. This is
is motivated by the fact that in Figure \ref{fig:motivation} (c) all curves are
above the diagonal gray line, i.e. a bounding box regressor trained for a
certain $u$ tends to produce bounding boxes of {\it higher\/} IoU. Hence,
starting from a set of examples $(x_i,\textbf{b}_i)$, cascade regression
successively resamples an example distribution $(x'_i,\textbf{b}'_i)$ of
higher IoU. In this manner, it is possible to keep the set of positive
examples of the successive stages at a roughly {\it constant\/} size,
even when the detector quality (IoU threshold) is {\it increased.\/} This is
illustrated in Figure \ref{fig:hist}, where the distribution tilts more
heavily towards high quality examples after each resampling step. Two
consequences ensue. First, there is no overfitting, since examples
are plentiful at all levels. Second, the detectors of the deeper
stages are optimized for higher IoU thresholds. Note that, some outliers
are sequentially removed by  increasing IoU thresholds, as illustrated in Figure \ref{fig:distribution}, enabling a better trained sequence of specialized detectors.

At each stage $t$, the R-CNN includes a classifier $h_t$ and a
regressor $f_t$ optimized for IoU threshold $u^t$, where $u^t>u^{t-1}$.
This is guaranteed by minimizing the loss
\begin{equation}
  L(x^t,g)=L_{cls}(h_t(x^t),y^t)+\lambda[y^t\geq{1}]L_{loc}(f_t(x^t,\textbf{b}^t),\textbf{g}),
\end{equation}
where $\textbf{b}^t=f_{t-1}(x^{t-1},\textbf{b}^{t-1})$, $g$ is the ground
truth object for $x^t$, $\lambda=1$ the trade-off coefficient, $[\cdot]$ the
indicator function, and $y^t$ is the label of $x^t$ given $u^t$
by (\ref{equ:cls label}). Unlike the \textit{integral loss} of
(\ref{equ:intloss}), this guarantees a sequence of effectively trained detectors of increasing quality. At inference, the quality of the hypotheses is sequentially improved, by applications of the same cascade procedure, and higher quality detectors are only required to operate on higher quality hypotheses. This enables high quality object detection, as suggested by Figure \ref{fig:motivation} (c) and (d).

\section{Experimental Results}

The Cascade R-CNN was evaluated on MS-COCO 2017 \cite{DBLP:conf/eccv/LinMBHPRDZ14}, which contains $\sim$118k images for training, 5k for validation (\texttt{val}) and $\sim$20k for testing without provided annotations (\texttt{test-dev}). The COCO-style Average Precision (AP) averages AP across IoU thresholds from 0.5 to 0.95 with an interval of 0.05. These evaluation metrics measure the detection performance of various qualities. All models were trained on COCO training set, and evaluated on \texttt{val} set. Final results were also reported on \texttt{test-dev} set.

\subsection{Implementation Details}

All regressors are class agnostic for simplicity. All cascade detection stages in Cascade R-CNN have the same architecture, which is the head of the baseline detection network. In total, Cascade R-CNN have four stages, one RPN and three for detection with $U=\{0.5,0.6,0.7\}$, unless otherwise noted. The sampling of the first detection stage follows \cite{DBLP:conf/iccv/Girshick15,DBLP:conf/nips/RenHGS15}. In the following stages, resampling is implemented by simply using
the regressed outputs from the previous stage, as in
Section \ref{subsec:cascade}. No data augmentation was used except standard horizontal image flipping. Inference was performed on a single image scale, with no further bells and whistles. All baseline detectors were reimplemented with Caffe \cite{DBLP:conf/mm/JiaSDKLGGD14}, on the same codebase for fair comparison.


\subsubsection{Baseline Networks}
\label{subsubsec:baseline}

To test the versatility of the Cascade R-CNN, experiments were performed with three popular baseline detectors: Faster-RCNN with backbone VGG-Net \cite{DBLP:journals/corr/SimonyanZ14a}, R-FCN \cite{DBLP:conf/nips/DaiLHS16} and FPN \cite{lin2017feature} with ResNet backbone \cite{DBLP:conf/cvpr/HeZRS16}. These baselines have a wide range of detection performances. Unless noted, their default settings were used. End-to-end training was used instead of multi-step training.

\paragraph{Faster-RCNN:} The network head has two fully connected layers. To reduce parameters, we used \cite{DBLP:conf/nips/HanPTD15} to prune less important connections. 2048 units were retained per fully connected layer and dropout layers were removed. Training started with a learning rate of 0.002, reduced by a factor of 10 at 60k and 90k iterations, and stopped at 100k iterations, on 2 synchronized GPUs, each holding 4 images per iteration. 128 RoIs were used per image.

\paragraph{R-FCN:} R-FCN adds a convolutional, a bounding box regression, and a classification layer to the ResNet. All heads of the Cascade R-CNN have this structure. Online hard negative mining \cite{DBLP:conf/cvpr/ShrivastavaGG16} was not used. Training started with a learning rate of 0.003, which was decreased by a factor of 10 at 160k and 240k iterations, and stopped at 280k iterations, on 4 synchronized GPUs, each holding one image per iteration. 256 RoIs were used per image.

\paragraph{FPN:} Since no source code is publicly available yet for FPN, our implementation details could be different. RoIAlign \cite{he2017mask} was used for a stronger baseline. This is denoted as FPN+ and was used in all ablation studies. As usual, ResNet-50 was used for ablation studies, and ResNet-101 for final detection. Training used a learning rate of 0.005 for 120k
iterations and 0.0005 for the next 60k iterations, on 8 synchronized GPUs,
each holding one image per iteration. 256 RoIs were used per image.

\begin{figure}[!t]
\begin{minipage}[b]{.48\linewidth}
\centering
\centerline{\epsfig{figure=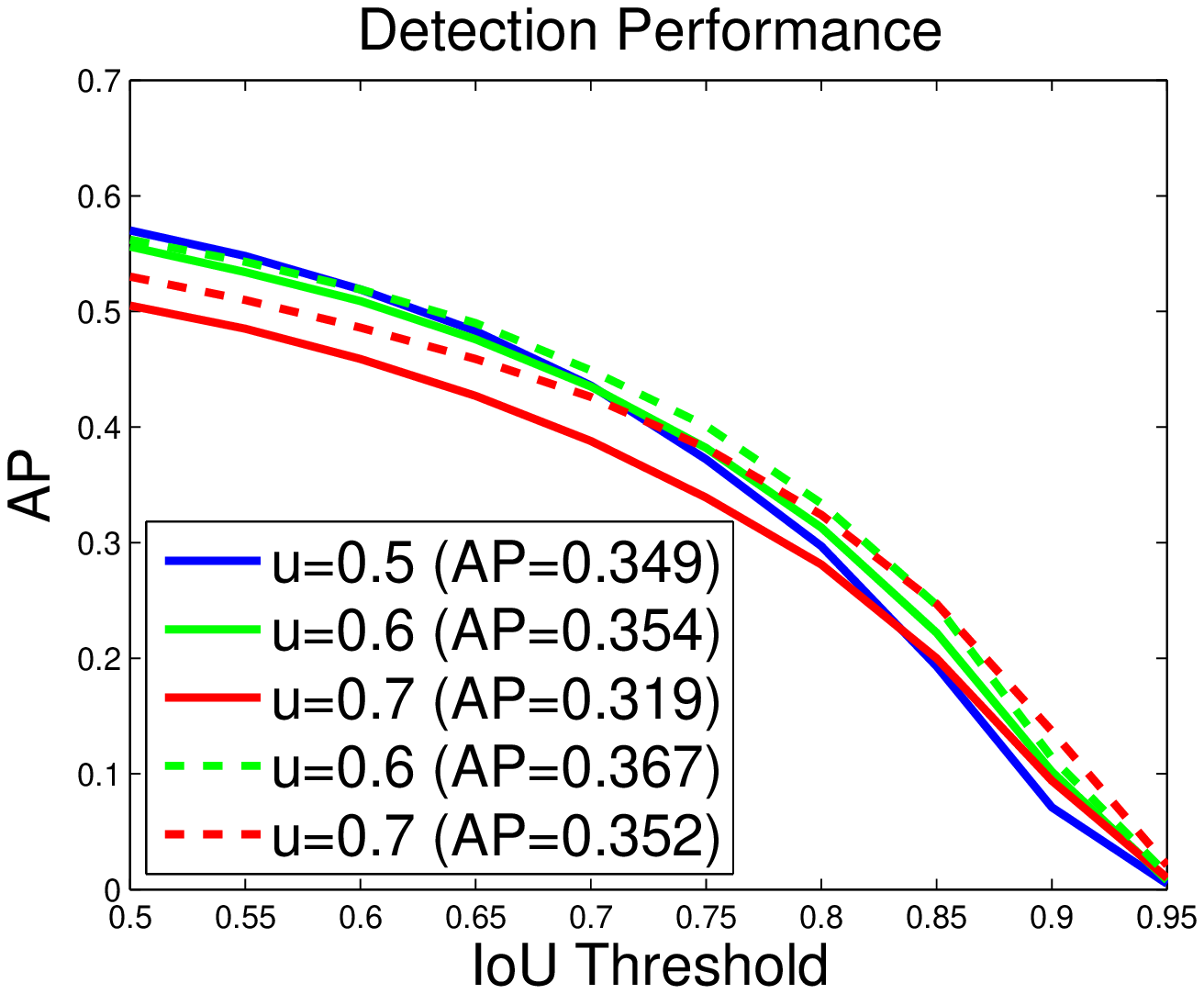,width=4.4cm,height=3.3cm}}{(a)}
\end{minipage}
\hfill
\begin{minipage}[b]{.48\linewidth}
\centering
\centerline{\epsfig{figure=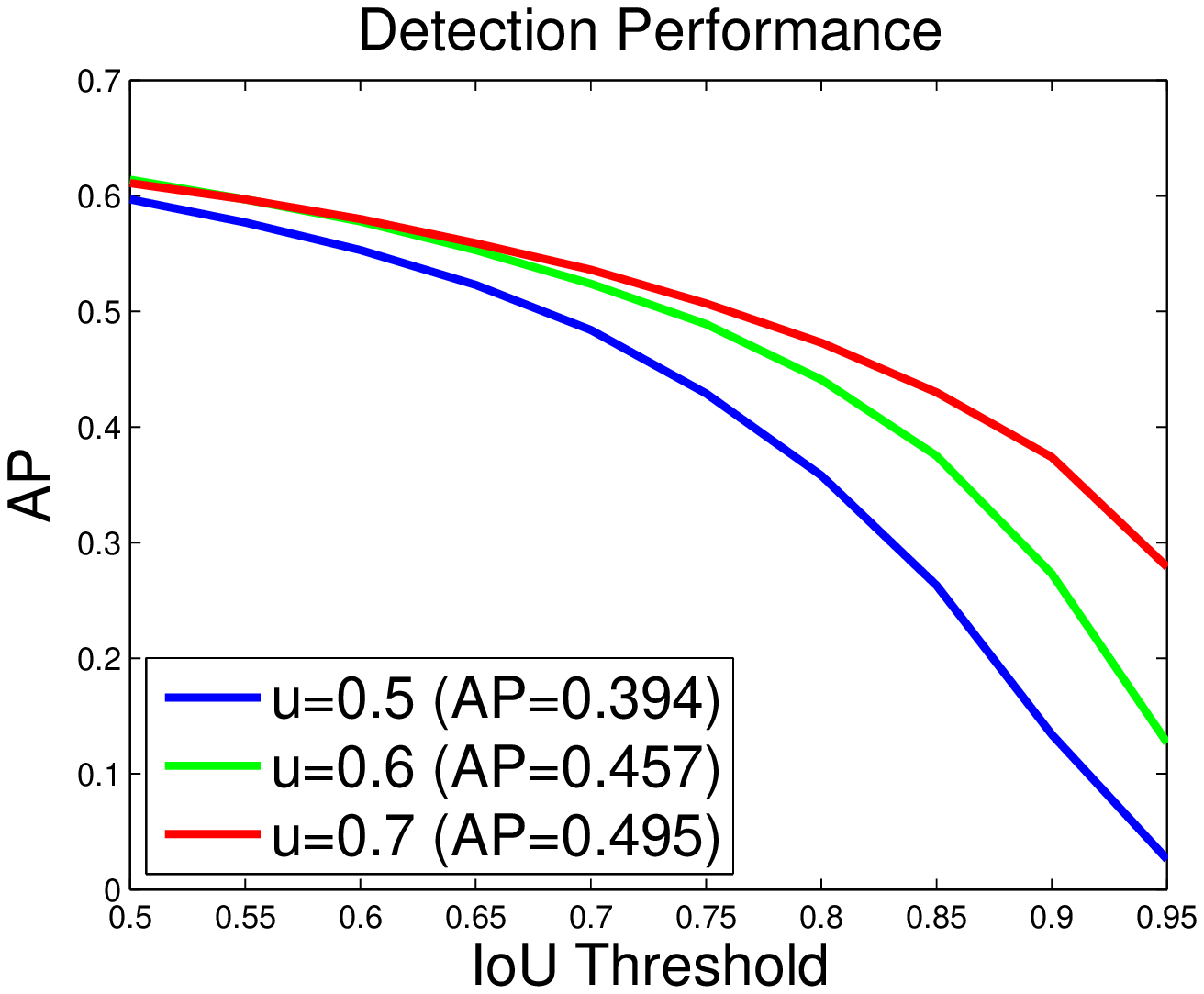,width=4.4cm,height=3.3cm}}{(b)}
\end{minipage}
\caption{(a) is detection performance of individually trained detectors, with their own proposals (solid curves) or Cascade R-CNN stage proposals (dashed curves), and (b) is by adding ground truth to the proposal set.}
\label{fig:hypotheses mismatch}
\end{figure}

\begin{figure}[!t]
\begin{minipage}[b]{.3\linewidth}
\centering
\centerline{\epsfig{figure=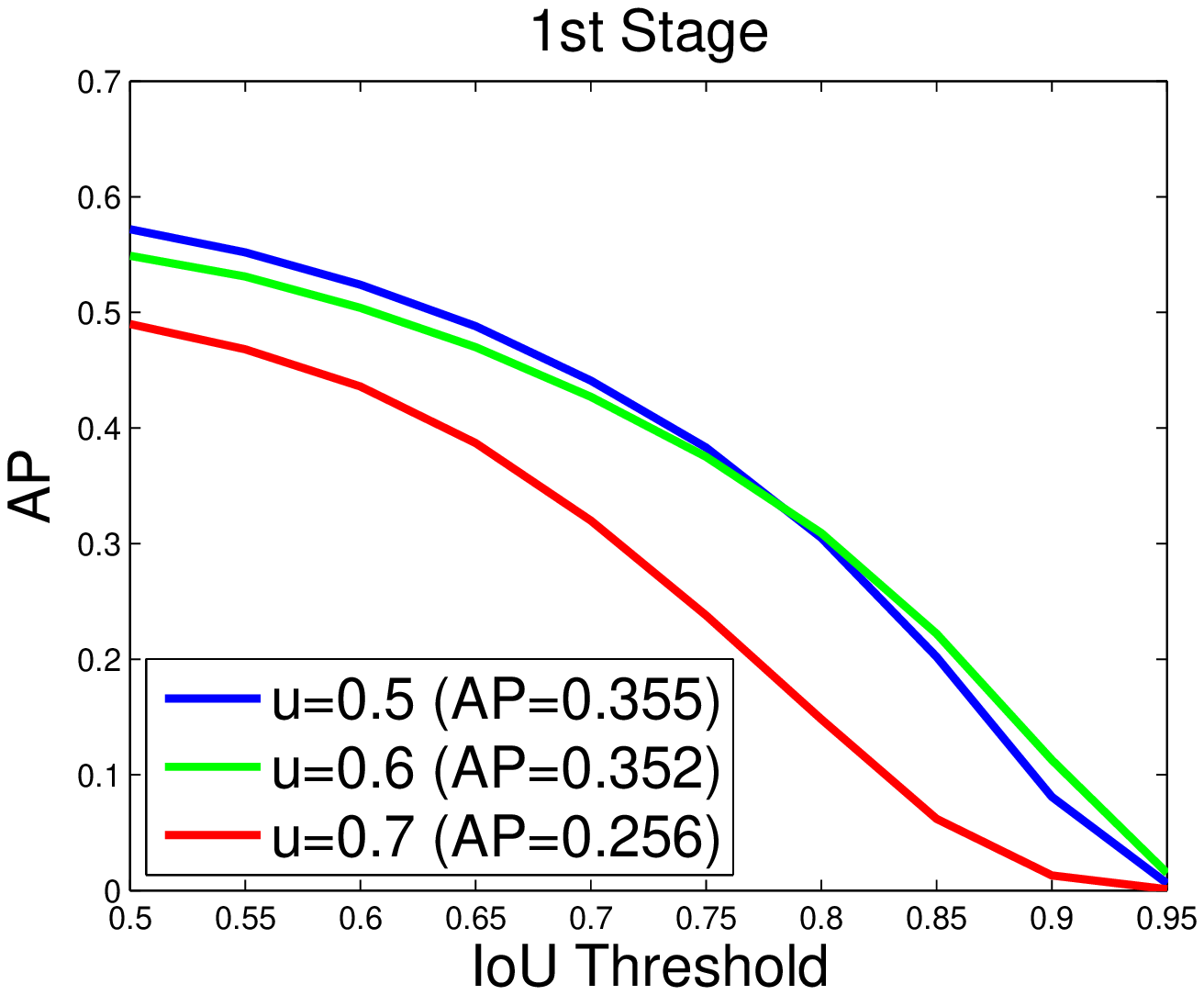,width=3.2cm,height=2.4cm}}
\end{minipage}
\hfill
\begin{minipage}[b]{.3\linewidth}
\centering
\centerline{\epsfig{figure=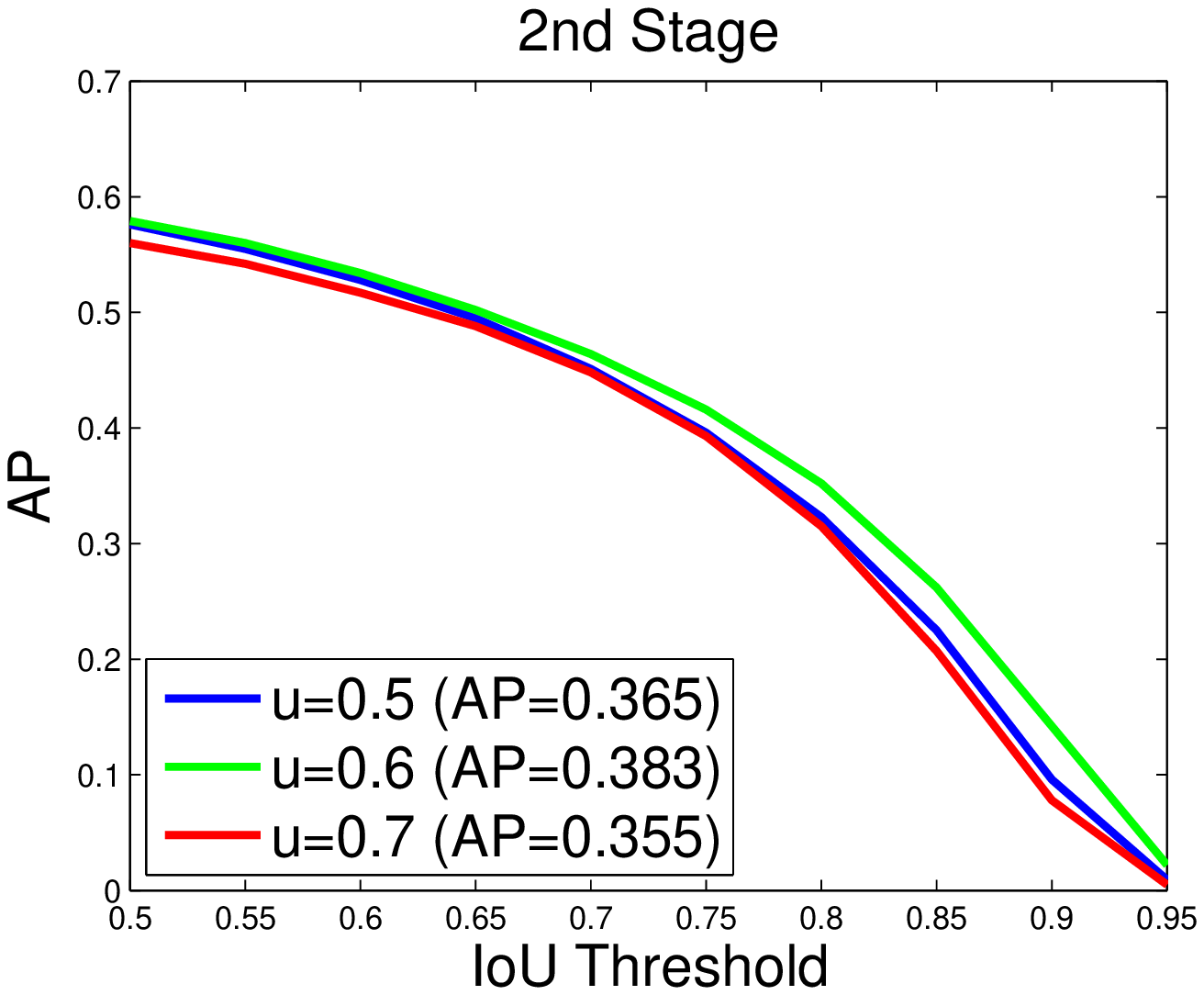,width=3.2cm,height=2.4cm}}
\end{minipage}
\hfill
\begin{minipage}[b]{.3\linewidth}
\centering
\centerline{\epsfig{figure=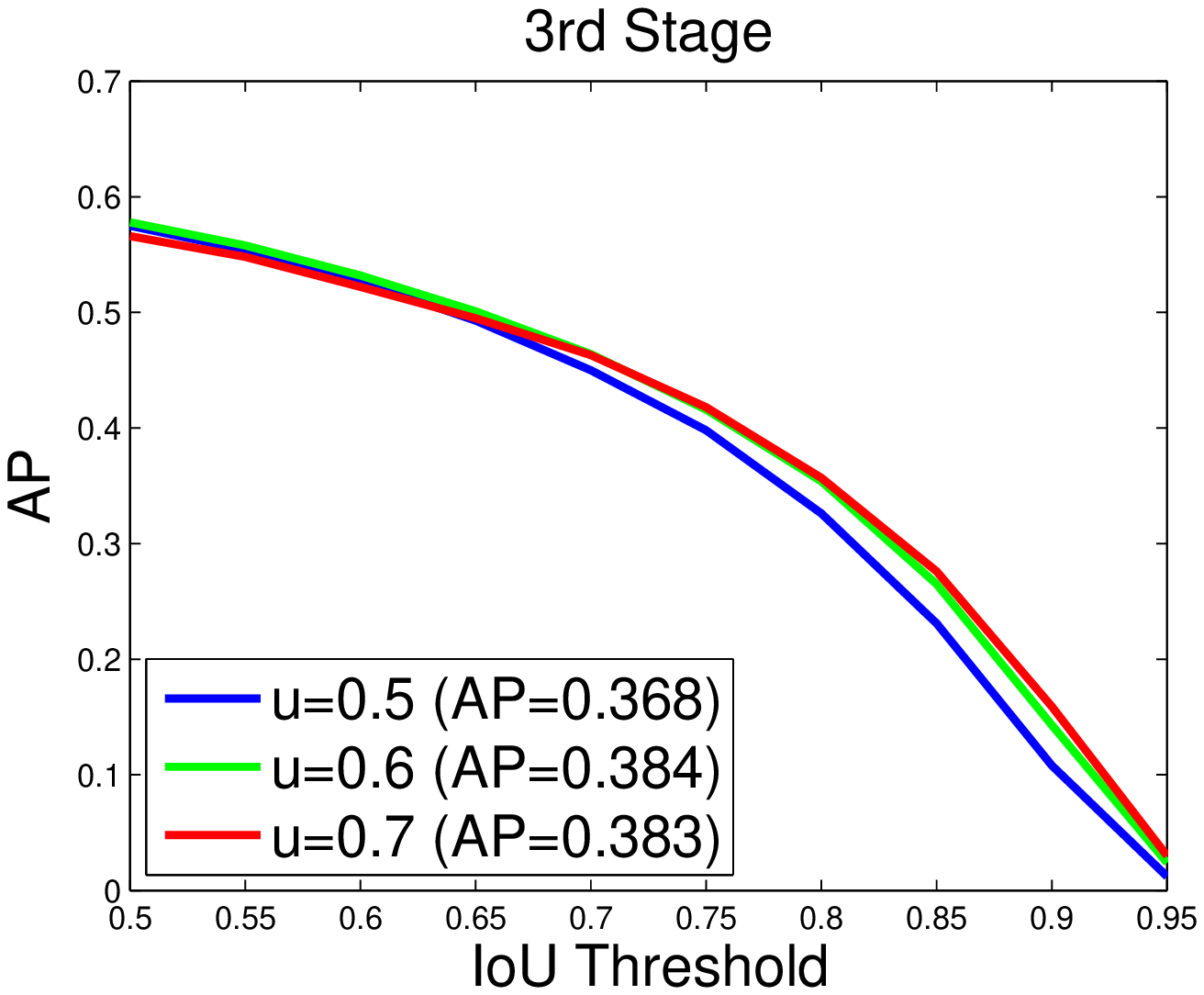,width=3.2cm,height=2.4cm}}
\end{minipage}
\caption{The detection performance of all Cascade R-CNN detectors at all cascade
stages.}
\label{fig:cascade stage}
\end{figure}

\subsection{Quality Mismatch}

Figure \ref{fig:hypotheses mismatch} (a) shows the AP curves of three
individually trained detectors of increasing IoU thresholds of
$U=\{0.5,0.6,0.7\}$. The detector of $u=0.5$ outperforms the detector
of $u=0.6$ at low IoU levels, but underperforms it at higher levels.
However, the detector of $u=0.7$ underperforms the other two.
To understand why this happens, we changed the quality of the proposals
at inference. Figure \ref{fig:hypotheses mismatch} (b)
shows the results obtained when ground truth bounding boxes were added to
the set of proposals. While all detectors improve, the detector of $u=0.7$
has the largest gains, achieving the best performance at almost all
IoU levels. These results suggest two conclusions. First, $u=0.5$ is not a
good choice for precise detection, simply more robust to low quality
proposals. Second, highly precise detection requires hypotheses that match
the detector quality. Next, the original detector proposals were replaced by the
Cascade R-CNN proposals of higher quality ($u=0.6$ and $u=0.7$ used the 2nd
and 3rd stage proposals, respectively). Figure \ref{fig:hypotheses mismatch} (a)
also suggests that the performance of the two detectors is significantly
improved when the testing proposals closer match the detector quality.

Testing all Cascade R-CNN detectors at all cascade stages produced
similar observations. Figure \ref{fig:cascade stage} shows that each detector was improved when used more precise hypotheses, while higher quality detector had larger gain. For example, the detector
of $u=0.7$ performed poorly for the low quality proposals of the 1st stage,
but much better for the more precise hypotheses available at the deeper
cascade stages. In addition, the jointly trained
detectors of Figure \ref{fig:cascade stage} outperformed the individually
trained detectors of Figure \ref{fig:hypotheses mismatch} (a), even
when the same proposals were used. This indicates that the detectors are
better trained within the Cascade R-CNN framework.

\begin{figure}[!t]
\begin{minipage}[b]{.48\linewidth}
\centering
\centerline{\epsfig{figure=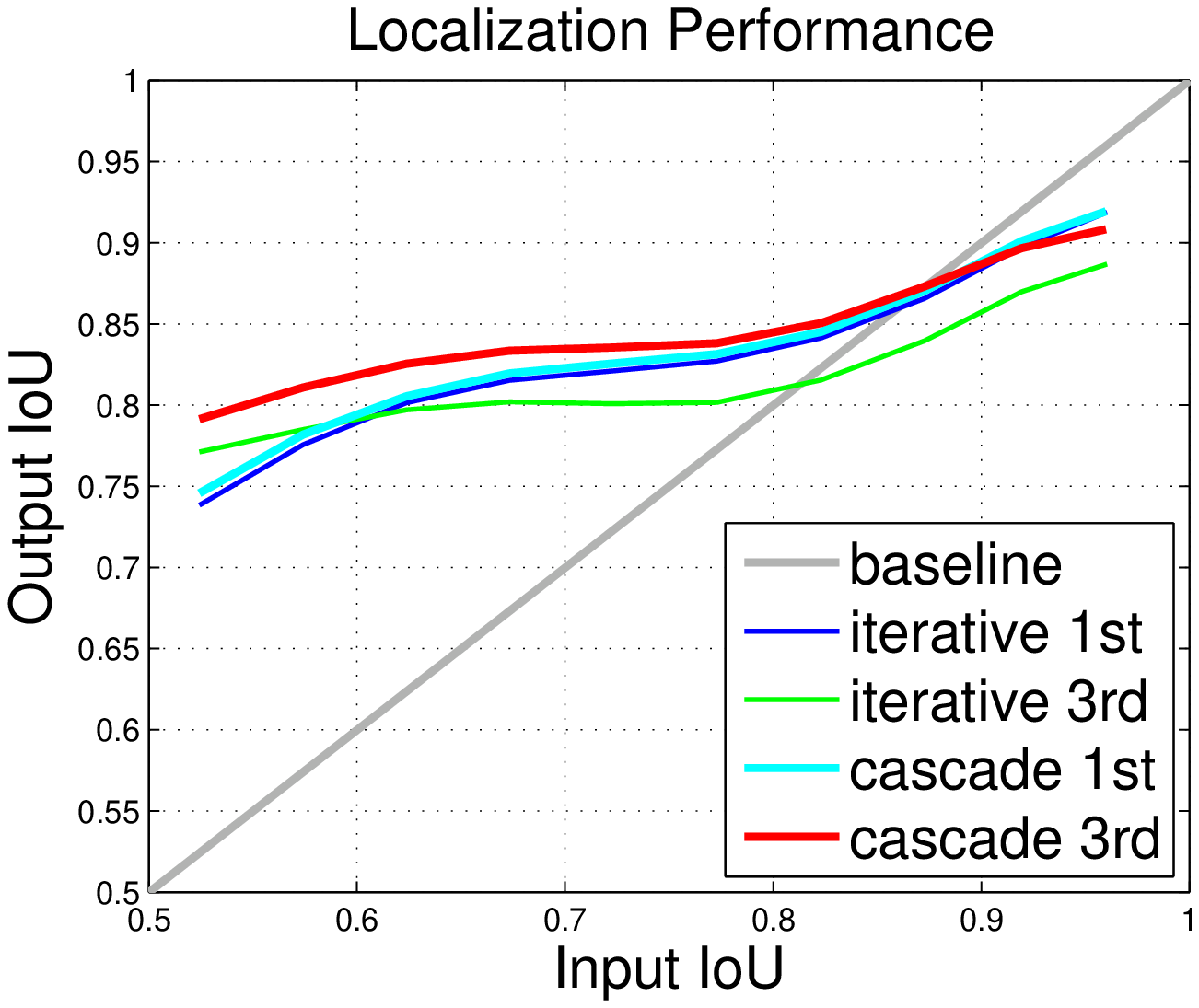,width=4.4cm,height=3.3cm}}{(a)}
\end{minipage}
\hfill
\begin{minipage}[b]{.48\linewidth}
\centering
\centerline{\epsfig{figure=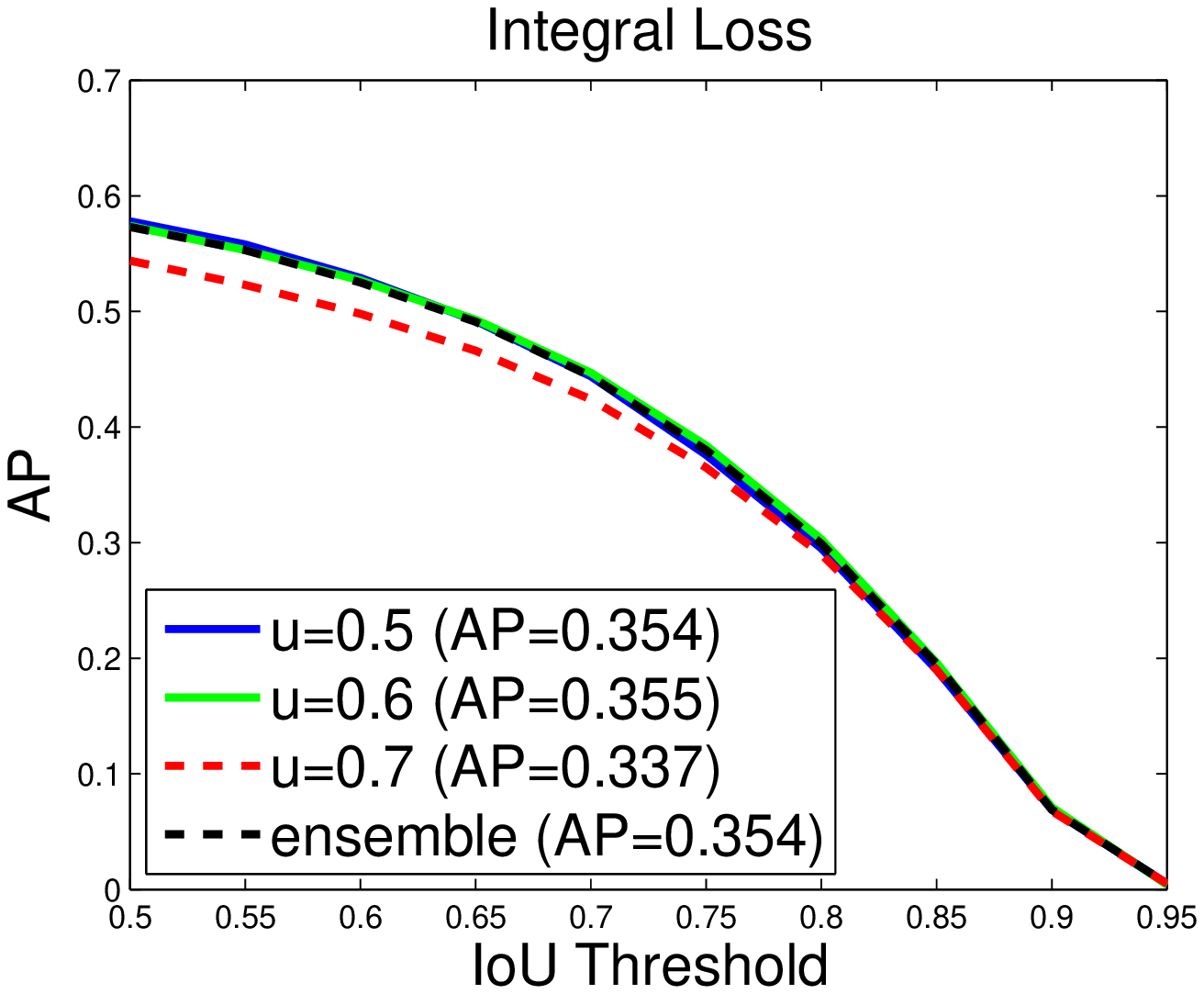,width=4.4cm,height=3.3cm}}{(b)}
\end{minipage}
\caption{(a) is the localization comparison, and (b) is the detection performance of individual classifiers in the \textit{integral loss} detector.}
\label{fig:localization and integral}
\end{figure}

\subsection{Comparison with \textit{Iterative BBox} and \textit{Integral Loss}}

In this section, we compare the Cascade R-CNN to \textit{iterative BBox} and the \textit{integral loss} detector. \textit{Iterative BBox} was implemented by applying the FPN+ baseline iteratively, three times. The \textit{integral loss} detector has the same number of classification heads as the Cascade R-CNN, with $U=\{0.5,0.6,0.7\}$.

\paragraph{Localization:} The localization performances of cascade regression
and \textit{iterative BBox} are compared in Figure \ref{fig:localization and integral} (a). The use of a single regressor degrades localization for hypotheses of high IoU. This effect accumulates when the regressor is applied iteratively, as in \textit{iterative BBox}, and performance actually drops. Note the very poor performance of \textit{iterative BBox} after 3 iterations. On the contrary, the cascade regressor has better performance at later stages, outperforming \textit{iterative BBox} at almost all IoU levels.

\begin{table}[t]
\tablestyle{1.8pt}{1.2}
\begin{tabular}{l|x{22}|x{22}x{22}x{22}x{22}x{22}}
& AP & AP$_{50}$ & AP$_{60}$ &AP$_{70}$ &AP$_{80}$ &AP$_{90}$\\ [.1em]
\shline
FPN+ baseline &34.9 &57.0  &51.9 &43.6 &29.7  &7.1\\
\textit{Iterative BBox} &35.4 &57.2  &52.1 &44.2 &30.4  &8.1\\
\textit{Integral Loss} &35.4 &57.3  &52.5 &44.4 &29.9  &6.9\\\hline
Cascade R-CNN &\bd{38.9} &\bd{57.8}  &\bd{53.4} &\bd{46.9} &\bd{35.8}  &\bd{15.8}\\
\end{tabular}\vspace{2mm}
\caption{The comparison with \textit{iterative BBox} and \textit{integral loss}.}
\label{tab:comparison}
\end{table}

\paragraph{Integral Loss:} The detection performances of all classifiers
in the \textit{integral loss} detector, sharing a single regressor, are shown in
Figure \ref{fig:localization and integral} (b). The classifier of $u=0.6$
is the best at all IoU levels, while the classifier of $u=0.7$ is the worst.
The ensemble of all classifiers shows no visible gain.

Table \ref{tab:comparison} shows, both \textit{iterative BBox} and \textit{integral loss} detector improve the baseline detector marginally. The cascade R-CNN has the best performance for all evaluation metrics. The gains are mild for low IoU thresholds but significant for the higher ones.

\begin{table}[t]
\tablestyle{1.8pt}{1.2}
\begin{tabular}{c|x{22}|x{22}x{22}x{22}x{22}x{22}}
test stage & AP & AP$_{50}$ & AP$_{60}$ &AP$_{70}$ &AP$_{80}$ &AP$_{90}$\\ [.1em]\shline
1 &35.5 &57.2  &52.4 &44.1 &30.5  &8.1\\
2 &38.3 &57.9  &53.4 &46.4 &35.2  &14.2\\
3 &38.3 &56.6  &52.2 &46.3 &35.7  &\bd{15.9}\\
$\overline{1\sim{2}}$ &38.5 &\bd{58.2}  &\bd{53.8} &46.7 &35.0  &14.0\\
$\overline{1\sim{3}}$ &\bd{38.9} &57.8  &53.4 &\bd{46.9} &\bd{35.8}  &15.8\\\hline
FPN+ baseline &34.9 &57.0  &51.9 &43.6 &29.7  &7.1\\
\end{tabular}\vspace{2mm}
\caption{The stage performance of Cascade R-CNN. $\overline{1\sim{3}}$ indicates the ensemble of three classifiers on the 3rd stage proposals.}
\label{tab:stage performance}
\end{table}

\subsection{Ablation Experiments}

Ablation experiments were also performed.

\paragraph{Stage-wise Comparison:} Table \ref{tab:stage performance} summarizes
stage performance. The 1st stage already outperforms the baseline detector,
due to the benefits of multi-stage multi-task learning. The 2nd stage improves
performance substantially, and the 3rd is equivalent to the 2nd. This
differs from the \textit{integral loss} detector, where the higher IOU
classifier is relatively weak. While the former (later) stage is better at low (high) IoU metrics, the ensemble of all classifiers is the best overall.

\paragraph{IoU Thresholds:} A preliminary Cascade R-CNN was trained using the same IoU threshold $u=0.5$ for all heads. In this case, the stages differ
only in the hypotheses they receive. Each stage is trained with the
corresponding hypotheses, i.e. accounting for the distributions of
Figure \ref{fig:distribution}. The first row of Table \ref{tab:ablation} shows
that the cascade improves on the baseline detector. This suggests the
importance of optimizing stages for the corresponding sample distributions.
The second row shows that, by increasing the stage threshold $u$, the
detector can be made more selective against close false positives and {\it specialized\/} for more precise hypotheses, leading to additional gains. This supports the conclusions of Section \ref{subsec:cascade}.

\paragraph{Regression Statistics:} Exploiting the progressively updated regression statistics, of Figure \ref{fig:distribution}, helps the effective multi-task learning of classification and regression. Its benefit is noted by comparing the models with/without it in Table \ref{tab:ablation}. The learning is not sensitive to these statistics.

\begin{table}[t]
\tablestyle{1.8pt}{1.2}
\begin{tabular}{cx{18}|x{22}|x{22}x{22}x{22}x{22}x{22}}
IoU$\uparrow$ & \textit{stat} & AP & AP$_{50}$ & AP$_{60}$ &AP$_{70}$ &AP$_{80}$ &AP$_{90}$\\ [.1em]\shline
& &36.8 &57.8  &52.9 &45.4 &32.0  &10.7\\
\cmark & &38.5 &\bd{58.4}  &\bd{54.1} &\bd{47.1} &35.0  &13.1\\
&\cmark &37.5 &57.8  &53.1 &45.5 &33.3  &13.1\\
\cmark &\cmark &\bd{38.9} &57.8  &53.4 &46.9 &\bd{35.8} &\bd{15.8}\\
\end{tabular}\vspace{2mm}
\caption{The ablation experiments. ``IoU$\uparrow$'' means increasing IoU thresholds, and ``\textit{stat}'' exploiting sequential regression statistics.}
\label{tab:ablation}
\end{table}

\begin{table}[t]
\tablestyle{1.8pt}{1.2}
\begin{tabular}{c|c|x{22}|x{22}x{22}x{22}x{22}x{22}}
\# stages &test stage & AP & AP$_{50}$ & AP$_{60}$ &AP$_{70}$ &AP$_{80}$ &AP$_{90}$\\ [.1em]\shline
1 &1 &34.9 &57.0  &51.9 &43.6 &29.7  &7.1\\
2 &$\overline{1\sim{2}}$ &38.2 &\bd{58.0}  &\bd{53.6} &46.7 &34.6  &13.6\\
3 &$\overline{1\sim{3}}$ &\bd{38.9} &57.8  &53.4 &\bd{46.9} &35.8  &15.8\\
4 &$\overline{1\sim{3}}$ &\bd{38.9} &57.4  &53.2 &46.8 &\bd{36.0}  &16.0\\
4 &$\overline{1\sim{4}}$ &38.6 &57.2  &52.8 &46.2 &35.5  &\bd{16.3}\\
\end{tabular}\vspace{2mm}
\caption{The impact of the number of stages in Cascade R-CNN.}
\label{tab:cascade stage}
\end{table}

\paragraph{Number of Stages:} The impact of the number of stages
is summarized in Table \ref{tab:cascade stage}. Adding a second
detection stage significantly improves the baseline detector.
Three detection stages still produce non-trivial improvement, but the
addition of a 4th stage ($u=0.75$) led to a slight performance
decrease. Note, however, that while the overall AP performance degrades,
the four-stage cascade has the best performance for high IoU levels. The three-stage cascade achieves the best trade-off.

\begin{table*}[t]
\tablestyle{3.5pt}{1.1}
\begin{tabular}{l|l|x{22}x{22}x{22}|x{22}x{22}x{22}}
&backbone &AP &AP$_{50}$ &AP$_{75}$  &AP$_{S}$ &AP$_{M}$ &AP$_{L}$\\\shline
YOLOv2 \cite{DBLP:conf/cvpr/RedmonDGF16}       &DarkNet-19 &21.6 &44.0 &19.2  &5.0 &22.4 &35.5\\
SSD513 \cite{DBLP:conf/eccv/LiuAESRFB16}    &ResNet-101 &31.2 &50.4 &33.3  &10.2 &34.5 &49.8\\
RetinaNet \cite{lin2017focal} &ResNet-101  &39.1 &59.1 &42.3  &21.8 &42.7 &50.2\\\hline
Faster R-CNN+++ \cite{DBLP:conf/cvpr/HeZRS16}*      &ResNet-101 &34.9 &55.7 &37.4  &15.6 &38.7 &50.9\\
Faster R-CNN w FPN \cite{lin2017feature}   &ResNet-101 &36.2 &59.1 &39.0  &18.2 &39.0 &48.2\\
Faster R-CNN w FPN+ (ours) &ResNet-101 &38.8 &61.1 &41.9  &21.3 &41.8 &49.8\\
Faster R-CNN by G-RMI \cite{DBLP:journals/corr/HuangRSZKFFWSG016} &Inception-ResNet-v2 &34.7 &55.5 &36.7  &13.5 &38.1 &52.0\\
Deformable R-FCN \cite{dai2017deformable}*   &Aligned-Inception-ResNet &37.5 &58.0 &40.8  &19.4 &40.1 &52.5\\
Mask R-CNN \cite{he2017mask}       &ResNet-101 &38.2 &60.3 &41.7  &20.1 &41.1 &50.2\\\hline
AttractioNet \cite{DBLP:conf/bmvc/GidarisK16}* &VGG16+Wide ResNet &35.7 &53.4 &39.3  &15.6 &38.0 &52.7\\
\bd{Cascade R-CNN} &ResNet-101 &\bd{42.8} &\bd{62.1} &\bd{46.3}  &\bd{23.7} &\bd{45.5} &\bd{55.2}\\\hline
\end{tabular}
\vspace{0.1cm}
\caption{Comparison with the state-of-the-art \emph{single-model} detectors on COCO \texttt{test-dev}. The entries denoted by ``*'' used bells and whistles at inference.}
\label{tab:state-of-the-art}
\end{table*}

\begin{table*}[t]
\tablestyle{1.8pt}{1.2}
\begin{tabular}{c|c|c|c|c|c|x{18}x{18}x{18}x{18}x{18}x{18}|x{18}x{18}x{18}x{18}x{18}x{18}}
& \multirow{2}{*}{backbone} & \multirow{2}{*}{cascade} & train & test & \multirow{2}{*}{param} &\multicolumn{6}{c|}{\texttt{val} (5k)} &\multicolumn{6}{c}{\texttt{test-dev} (20k)}\\\cline{7-18}
& & &speed &speed & &AP & AP$_{50}$ & AP$_{75}$ &AP$_{S}$ &AP$_{M}$ &AP$_{L}$ & AP & AP$_{50}$ & AP$_{75}$ &AP$_{S}$ &AP$_{M}$ &AP$_{L}$\\ [.1em]
\shline
\multirow{2}{*}{Faster R-CNN} &\multirow{2}{*}{VGG} & \xmark &0.12s & 0.075s &278M &23.6 &43.9  &23.0 &8.0 &26.2  &35.5 &23.5 &43.9  &22.6 &8.1 &25.1  &34.7\\
& & \cmark &0.14s & 0.115s &704M &27.0 &44.2  &27.7 &8.6 &29.1  &42.2 &26.9 &44.3  &27.8 &8.3 &28.2  &41.1\\\hline
\multirow{2}{*}{R-FCN} &\multirow{2}{*}{ResNet-50} & \xmark &0.19s & 0.07s &133M &27.0 &48.7  &26.9 &9.8 &30.9  &40.3 &27.1 &49.0  &26.9 &10.4 &29.7  &39.2\\
& & \cmark &0.24s & 0.075s &184M &31.1 &49.8  &32.8 &10.4 &34.4  &48.5 &30.9 &49.9  &32.6 &10.5 &33.1  &46.9\\\hline
\multirow{2}{*}{R-FCN} &\multirow{2}{*}{ResNet-101} & \xmark &0.23s & 0.075s &206M &30.3 &52.2  &30.8 &12.0 &34.7  &44.3 &30.5 &52.9  &31.2 &12.0 &33.9  &43.8\\
& & \cmark &0.29s & 0.083s &256M &33.3 &52.0  &35.2 &11.8 &37.2  &51.1 &33.3 &52.6  &35.2 &12.1 &36.2  &49.3\\\hline
\multirow{2}{*}{FPN+} &\multirow{2}{*}{ResNet-50} & \xmark &0.30s & 0.095s &165M &36.5 &58.6  &39.2 &20.8 &40.0  &47.8 &36.5 &59.0  &39.2 &20.3 &38.8  &46.4\\
& & \cmark &0.33s & 0.115s &272M &40.3 &59.4  &43.7 &22.9 &43.7  &54.1 &40.6 &59.9  &44.0 &22.6 &42.7  &52.1\\\hline
\multirow{2}{*}{FPN+} &\multirow{2}{*}{ResNet-101} & \xmark &0.38s & 0.115s &238M &38.5 &60.6  &41.7 &22.1 &41.9  &51.1 &38.8 &61.1 &41.9  &21.3 &41.8 &49.8\\
& & \cmark &0.41s & 0.14s &345M &42.7 &61.6  &46.6 &23.8 &46.2  &57.4 &42.8 &62.1 &46.3  &23.7 &45.5 &55.2\\\hline
\end{tabular}\vspace{2mm}
\caption{Detailed comparison on multiple popular baseline object detectors. All speeds are reported per image on a single Titan Xp GPU.}
\label{tab:generalization}
\end{table*}

\subsection{Comparison with the state-of-the-art}
\label{subsec:state-of-the-art}

The Cascade R-CNN, based on FPN+ and ResNet-101 backbone, is compared to state-of-the-art \emph{single-model} object detectors in Table \ref{tab:state-of-the-art}. The settings are as described in
Section \ref{subsubsec:baseline}, but a total of 280k training
iterations were run and the learning rate dropped at 160k and 240k
iterations. The number of RoIs was also increased to 512. The first group
of detectors on Table \ref{tab:state-of-the-art} are one-stage detectors,
the second group two-stage, and the last group multi-stage
(3-stages+RPN for the Cascade R-CNN). All the compared state-of-the-art detectors were trained with $u=0.5$. It is noted that our FPN+ implementation is
better than the original FPN \cite{lin2017feature}, providing a very strong
baseline. In addition, the extension from FPN+ to Cascade R-CNN improved
performance by $\sim$4 points. The Cascade R-CNN also outperformed
all \emph{single-model} detectors by a large margin, under all evaluation metrics. This includes the \emph{single-model} entries of the COCO challenge winners in 2015 and 2016 (Faster R-CNN+++ \cite{DBLP:conf/cvpr/HeZRS16},
and G-RMI \cite{DBLP:journals/corr/HuangRSZKFFWSG016}), and the very recent
Deformable R-FCN \cite{dai2017deformable}, RetinaNet \cite{lin2017focal}
and Mask R-CNN \cite{he2017mask}. The best multi-stage detector on COCO,
AttractioNet \cite{DBLP:conf/bmvc/GidarisK16}, used \textit{iterative BBox} for proposal generation. Although many enhancements were used in AttractioNet, the vanilla Cascade R-CNN still outperforms it by 7.1 points. Note that, unlike Mask R-CNN, no segmentation information is exploited in the Cascade R-CNN. Finally, the vanilla \emph{single-model} Cascade R-CNN also surpasses the heavily engineered ensemble detectors that won the COCO challenge in 2015 and 2016 (AP 37.4 and 41.6, respectively)\footnote{http://cocodataset.org/\#detections-leaderboard}.

\subsection{Generalization Capacity}

Three-stage Cascade R-CNN of all three baseline detectors
are compared in Table \ref{tab:generalization}. All settings are as above,
with the changes of Section \ref{subsec:state-of-the-art} for FPN+.

\paragraph{Detection Performance:} Again, our implementations are better than
the original detectors \cite{DBLP:conf/nips/RenHGS15,DBLP:conf/nips/DaiLHS16,lin2017feature}. Still, the Cascade R-CNN improves on these baselines
consistently by 2$\sim$4 points, independently of their strength.
These gains are also consistent on \texttt{val} and \texttt{test-dev}. These
results suggest that the Cascade R-CNN is widely applicable across
detector architectures.

\paragraph{Parameter and Timing:} The number of the Cascade R-CNN parameters increases with the number of cascade stages. The increase is linear in the parameter number of the baseline detector heads. In addition, because the computational cost of a detection head is usually small when compared to the RPN, the computational overhead of the Cascade R-CNN is small, at both training and testing.

\section{Conclusion}

In this paper, we proposed a multi-stage object detection framework,
the Cascade R-CNN, for the design of high quality object detectors.
This architecture was shown to avoid the problems of overfitting at
training and quality mismatch at inference. The solid and consistent
detection improvements of the Cascade R-CNN on the challenging COCO dataset
suggest the modeling and understanding of various concurring factors
are required to advance object detection. The Cascade R-CNN was shown to be
applicable to many object detection architectures. We believe that it can be
useful to many future object detection research efforts.

\paragraph{Acknowledgment} We would like to thank Kaiming He for valuable discussions.

{\small
\bibliographystyle{ieee}
\bibliography{egbib}
}

\end{document}